\documentclass[twoside,11pt]{article}

\usepackage{amsmath}
\usepackage{amssymb}
\usepackage{comment}
\usepackage{graphicx}
\usepackage{cancel}
\usepackage{natbib}

\usepackage{dsfont}
\usepackage{xcolor}

\usepackage{ stmaryrd }
\usepackage{enumitem} 

\usepackage{setspace}
\usepackage{enumitem}
\usepackage{wrapfig} 
\usepackage[normalem]{ulem}

\usepackage{times}
\usepackage{graphicx} 
\usepackage{subfigure} 
\graphicspath{{./},{./figures/}}

\usepackage[textsize=small]{todonotes}

\usepackage[preprint]{jmlr2e}

\usepackage{cleveref}
\newtheorem{assumption}{\textbf{A}\hspace{-3pt}}
\Crefname{assumption}{\textbf{A}\hspace{-3pt}}{\textbf{A}\hspace{-3pt}}
\crefname{assumption}{\textbf{A}}{\textbf{A}}

\jmlrheading{1}{2019}{1-32}{11/19}{--/--}{tbd}{\c{S}im\c{s}ekli et al.}

\ShortHeadings{Heavy-Tailed Theory of Stochastic Gradient Descent}{Simsekli et al.\ }  
\firstpageno{1}

\newcommand{\sil}[1]{}

\newcommand{\wb}{\mathbf{w}}
\newcommand{\rset}{\mathbb{R}}
\newcommand{\E}{\mathbb{E}}

\newcommand{\rmd}{\mathrm{d}}
\newcommand{\Bm}{\mathrm{B}}
\newcommand{\Lm}{\mathrm{L}^\alpha}
\newcommand{\Id}{\mathbf{I}}
\newcommand{\beq}{\begin{eqnarray}}
\newcommand{\eeq}{\end{eqnarray}}
\newcommand{\beqs}{\begin{eqnarray*}}
\newcommand{\eeqs}{\end{eqnarray*}}

\newcommand{\sas}{{\cal S}\alpha{\cal S}}

\DeclareMathOperator*{\argmin}{arg\min}

\begin{document}

\title{On the Heavy-Tailed Theory of Stochastic Gradient Descent for Deep Neural Networks}

\author{\name Umut \c{S}im\c{s}ekli$^\text{*1,2}$ \email umut.simsekli@telecom-paris.fr 
       \AND
      \name Mert G\"{u}rb\"{u}zbalaban$^\text{*3}$ \email mg1366@rutgers.edu 
      \AND
      \name Thanh Huy Nguyen$^\text{1}$ \email thanh.nguyen@telecom-paris.fr 
      \AND
      \name Ga\"{e}l Richard$^\text{1}$ \email gael.richard@telecom-paris.fr 
      \AND
      \name Levent Sagun$^\text{4}$ \email leventsagun@fb.com 
      \AND
      \addr 1: LTCI, T\'{e}l\'{e}com Paris, Institut Polytechnique de Paris, Paris, France\\
      2: Department of Statistics, University of Oxford, Oxford, UK \\
      3: Rutgers Business School, NJ, USA \\
      4: Facebook AI Research, Paris, France \\
      * denotes equal contribution
      }

\editor{TBD}

\maketitle

\begin{abstract}
The gradient noise (GN) in the stochastic gradient descent (SGD) algorithm is often considered to be Gaussian in the large data regime by assuming that the \emph{classical} central limit theorem (CLT) kicks in. 
This assumption is often made for mathematical convenience, since it enables SGD to be analyzed as a stochastic differential equation (SDE) driven by a Brownian motion.
We argue that the Gaussianity assumption might fail to hold in deep learning settings and hence render the Brownian motion-based analyses inappropriate. Inspired by non-Gaussian natural phenomena, we consider the GN in a more general context and invoke the \emph{generalized} CLT, which suggests that the GN converges to a \emph{heavy-tailed} $\alpha$-stable random vector, where \emph{tail-index} $\alpha$ determines the heavy-tailedness of the distribution.
Accordingly, we propose to analyze SGD as a discretization of an SDE driven by a L\'{e}vy motion. Such SDEs can incur `jumps', which force the SDE and its discretization \emph{transition} from narrow minima to wider minima, as proven by existing metastability theory and the extensions that we proved recently. In this study, under the $\alpha$-stable GN assumption, we further establish an explicit connection between the convergence rate of SGD to a local minimum and the tail-index $\alpha$. 
To validate the $\alpha$-stable assumption, we conduct experiments on common deep learning scenarios and show that in all settings, the GN is highly non-Gaussian and admits heavy-tails. We investigate the tail behavior in varying network architectures and sizes, loss functions, and datasets. Our results open up a different perspective and shed more light on the belief that SGD prefers wide minima.
\end{abstract}


\tableofcontents

\section{Introduction}
\label{sec:intro}

\subsection{Context and motivation}

Deep neural networks have revolutionized machine learning and have ubiquitous use in many application domains \citep{hinton-nature,Krizhevsky12,Hinton12}. 
In full generality, many key tasks in deep learning reduce to solving the following optimization problem:
\begin{align}
\wb^\star = \argmin_{\wb \in \rset^d} \Bigl\{ f(\wb) \triangleq \frac1{n} \sum_{i=1}^n f^{(i)}(\wb) \Bigr\}
\end{align}
where $\wb\in\rset^d$ denotes the weights of the neural network, $f:\rset^d\to\rset$ denotes the loss function that is typically non-convex in $\wb$, each $f^{(i)}$ denotes the (instantaneous) loss function that is contributed by the \emph{data point} $i \in \{1, \dots, n\}$, and $n$ denotes the total number of data points.
Stochastic gradient descent (SGD) is one the most popular approaches for attacking this problem in practice and is based on the following iterative updates:
\begin{align}
 \wb^{k+1} = \wb^{k} - \eta \nabla \tilde{f}_{k}(\wb^k) \label{eqn:sgd_main}
\end{align} 
where $k \in \{1, \dots, K\}$ denotes the iteration number, $\eta$ is the step-size (or the learning rate), and $\nabla \tilde{f}_{k}$ denotes the stochastic gradient at iteration $k$, that is defined as follows:
\begin{align}
\label{eqn:stoch_grad}
\nabla \tilde{f}_{k} (\wb) \triangleq \nabla \tilde{f}_{\Omega_k} (\wb) \triangleq \frac1{b} \sum_{i \in \Omega_k}  \nabla f^{(i)}(\wb).
\end{align} 
Here, $\Omega_k \subset \{1,\dots,n\}$ is a random subset that is drawn with or without replacement at iteration $k$, and $b = |\Omega_k|$ denotes the number of elements in $\Omega_k$.

SGD is widely used in deep learning with a great success in its computational efficiency \citep{bottou2010large,bottou2008tradeoffs,pmlr-v80-daneshmand18a}. Beyond efficiency, understanding how SGD performs better than its full batch counterpart in terms of test accuracy remains a major challenge. Even though SGD seems to find perfect training performance at (near-) zero loss solutions on the training landscape (at least in certain regimes \citep{Zhang16, sagun2014explorations, keskar2016large, Geiger18}), it appears that the algorithm finds solutions with different properties depending on how it is tuned \citep{sutskever2013importance, keskar2016large, jastrzkebski2017three, hoffer2017train, masters2018revisiting, smith2017don}. Despite the fact that the impact of SGD on generalization has been studied \citep{advani2017high, wu2018sgd, neyshabur2017exploring}, a satisfactory theory that can explain its success in a way that encompasses such peculiar empirical properties is still lacking.

A popular approach for investigating the behavior of SGD is based on considering SGD as a discretization of a continuous-time process \citep{mandt2016variational,jastrzkebski2017three,pmlr-v70-li17f,hu2017diffusion,zhu2018anisotropic,chaudhari2018stochastic}. This approach models the stochastic gradient noise as a Gaussian distribution, i.e. $U_k(\wb) \triangleq \nabla \tilde{f}_{k} (\wb) - \nabla f(\wb)$ satisfies
\begin{align}
U_k(\wb) \sim {\cal N}(\mathbf{0}, \sigma^2 \Id), \label{eqn:noise_gauss}
\end{align}
where ${\cal N}$ denotes the multivariate (Gaussian) normal distribution and $\Id$ denotes the identity matrix of appropriate size.\footnote{We note that more sophisticated assumptions than \eqref{eqn:noise_gauss} have been made in terms of the covariance matrix of the Gaussian distribution (e.g.\ state dependent, anisotropic). However, in all these cases, the resulting distribution is still a Gaussian, therefore the same criticism holds.} The rationale behind this assumption is that, if the size of the minibatch $b$ is large enough, then we can invoke the Central Limit Theorem (CLT) and assume that the distribution of $U_k$ is approximately Gaussian. Then, under this assumption, \eqref{eqn:sgd_main} can be written as follows:
\begin{align}
\wb^{k+1} = \wb^{k} - \eta \nabla f(\wb^k) + \sqrt{\eta} \sqrt{\eta \sigma^2} Z_k, \label{eqn:sgd_gauss}
\end{align}
where $Z_k$ denotes a standard normal random vector in $\rset^d$. If we further assume that $\eta$ is small enough, then the continuous-time analogue of the discrete-time process \eqref{eqn:sgd_gauss} is the following stochastic differential equation (SDE):
\begin{align}
\rmd \wb_t = - \nabla f(\wb_t) \rmd t + \sqrt{\eta \sigma^2} \rmd \Bm_t , \label{eqn:sgd_langevin}
\end{align}
where $\Bm_t$ denotes the standard Brownian motion. This SDE is a variant of the well-known \emph{Langevin diffusion} and under mild regularity assumptions on $f$, one can show that the Markov process $(\wb_t)_{t\geq 0}$ is ergodic with its unique invariant measure, whose density is proportional to $\exp(-f(x)/(\eta \sigma^2))$ for any $\eta>0$ \citep{Roberts03}. From this perspective, the SGD recursion in \eqref{eqn:sgd_gauss} can be seen as a first-order Euler-Maruyama discretization of the Langevin dynamics (see also \citep{pmlr-v70-li17f,jastrzkebski2017three,hu2017diffusion}), which is often referred to as the Unadjusted Langevin Algorithm (ULA) \citep{Roberts03,lamberton2003recursive,durmus2015non,durmus2016stochastic}.

Based on this observation, \citet{jastrzkebski2017three} focused on the relation between this invariant measure and the algorithm parameters, namely the step-size $\eta$ and mini-batch size, as a function of $\sigma^2$. They concluded that the ratio of step-size divided by the batch size is the control parameter that determines the width of the minima found by SGD. Furthermore, they revisit the famous wide minima folklore \citep{hochreiter1997flat}: Among the minima found by SGD, the wider it is, the better it performs on the test set. However, there are several fundamental issues with this approach, which we will explain below.

\begin{figure}[t]
    \centering
    \subfigure[Gradient noise (first)]{\includegraphics[width=0.23\columnwidth]{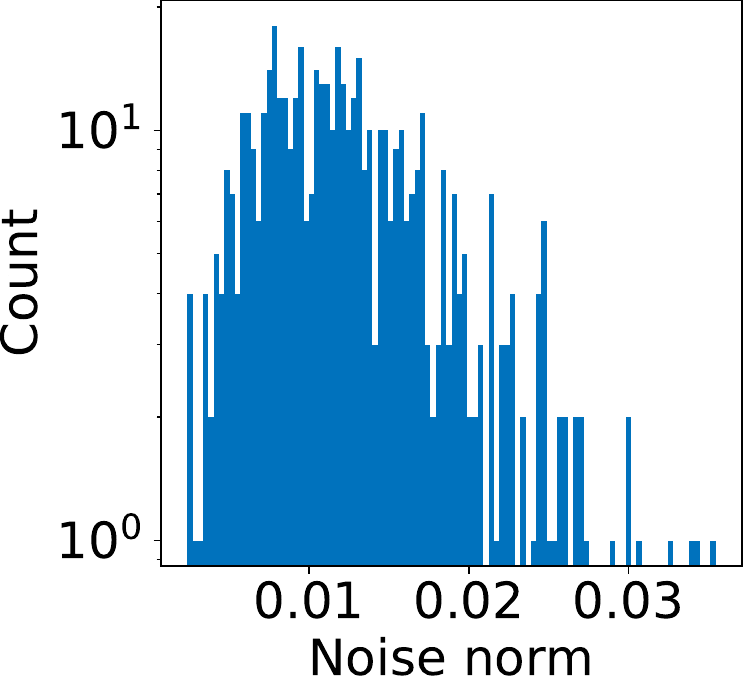}}
    \subfigure[Gradient noise (last)]{\includegraphics[width=0.23\columnwidth]{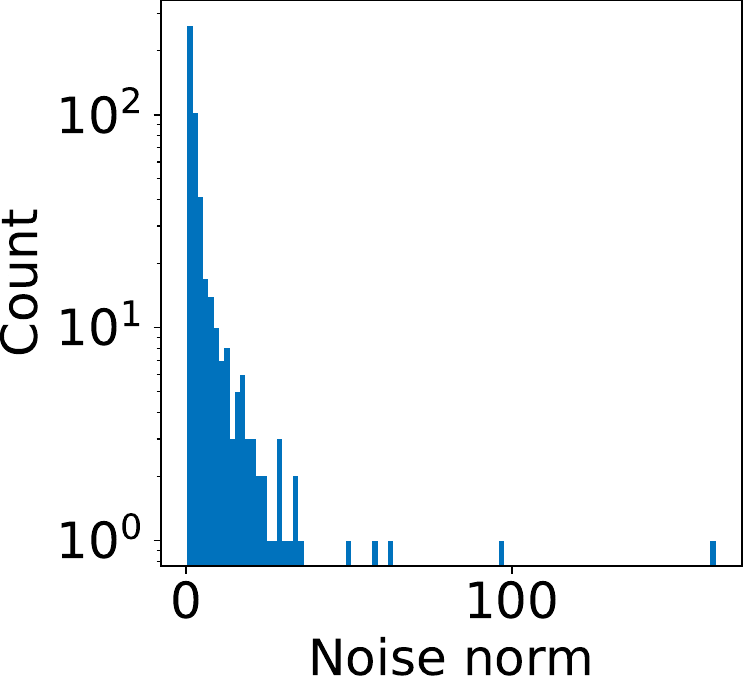}}
    \subfigure[Gaussian]{\includegraphics[width=0.23\columnwidth]{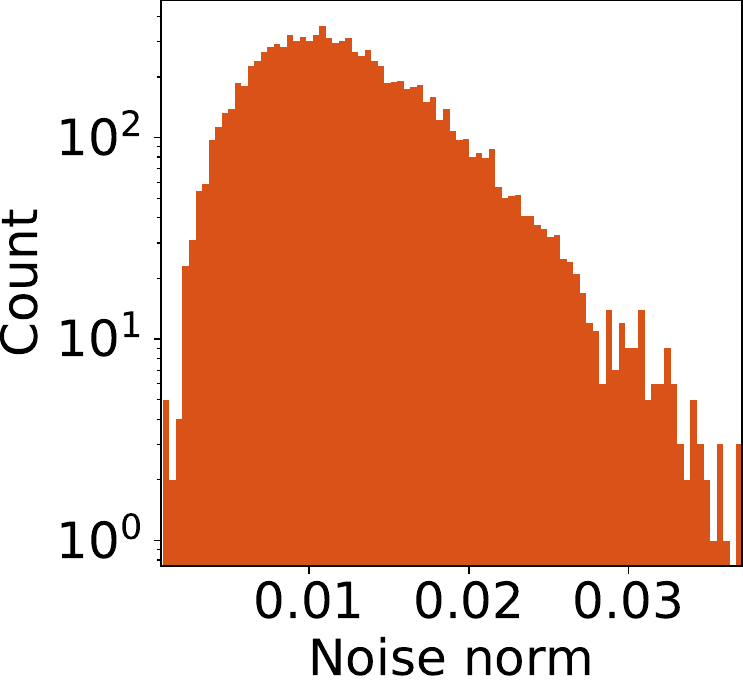}}
    \subfigure[$\alpha$-stable]{\includegraphics[width=0.23\columnwidth]{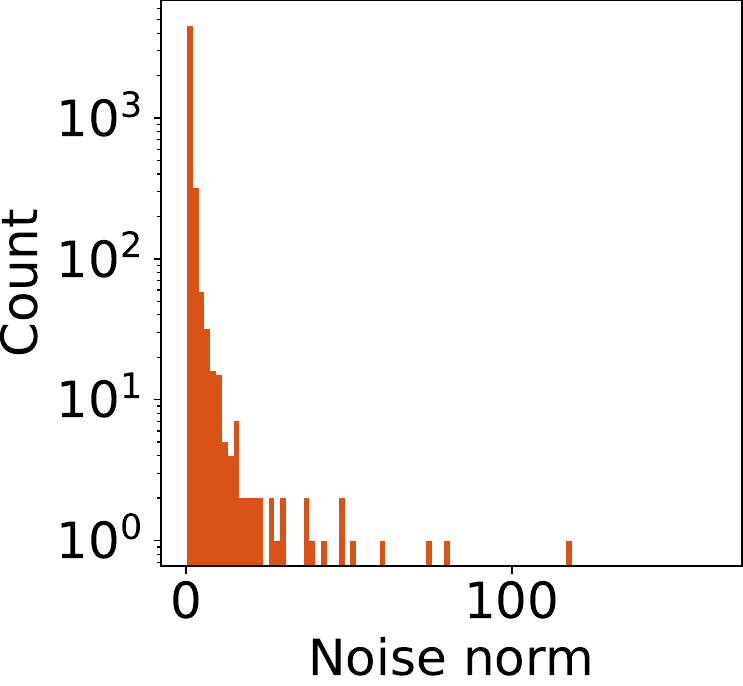} \label{fig:hist_sas}}
    \vspace{-5pt}
    \caption{(a) and (b) The histogram of the norms of the gradient noises after the first and the last iterations, respectively (computed with AlexNet on CIFAR10). (c) and (d) the histograms of the norms of (scaled) Gaussian and $\alpha$-stable random vectors, respectively. }
    \vspace{-10pt}
    \label{fig:noise_norms}
\end{figure}

We first illustrate a typical mismatch between the Gaussianity assumption and the empirical behavior of the stochastic gradient noise in terms of the long term behavior.
In Figure~\ref{fig:noise_norms}, we plot the histogram of the norms of the stochastic gradient noise at the first and the last iterations that are computed using a convolutional neural network (AlexNet) in an image classification problem on the CIFAR10 dataset and compare it to the histogram of the norms of Gaussian random vectors\footnote{In our conference proceeding \citep{simsekli_tail_ICML2019}, we have identified an error in the corresponding figure: the histogram of the norms was computed over \emph{all the gradients noises} that were obtained throughout training. This issue is fixed in this article.}. It can be clearly observed that, even though the shape of the histogram {corresponding to gradients} resembles the one of the Gaussian vectors at the first iteration, throughout training, it drifts apart from the Gaussian and exhibits a \emph{heavy-tailed} behavior.

In addition to the empirical observations, the Gaussianity assumption also yields some theoretical issues. The first issue with this assumption is that the current SDE analyses of SGD are based on the \emph{invariant measure} of the SDE, which implicitly assumes that sufficiently many iterations have been taken to converge to that measure. Recent results on ULA \citep{raginsky17a,xu2018global} have shown that, the required number of iterations to achieve the invariant measure often grows exponentially with the dimension $d$. This result contradicts with the current practice: considering the large size of the neural networks and limited computational budget, only a limited number of iterations -- which is much smaller than $\exp(\mathcal{O}(d))$ -- can be taken. This conflict becomes clearer in the light of the recent works that studied the \emph{local} behavior of ULA \citep{tzen2018local,zhang17b}. These studies showed that ULA will get close to the nearest local optimum in polynomial time; however, the required amount of time for escaping from that local optimum increases exponentially with the dimension. Therefore, the phenomenon that SGD prefers wide minima within a considerably small number of iterations cannot be explained using the asymptotic distribution of the SDE given in \eqref{eqn:sgd_langevin}.

The second issue is related to the local behavior of the process and becomes clear when we consider the \emph{metastability} analysis of Brownian motion-driven SDEs.  These studies \citep{freidlin1998random,bovier2004metastability,Imkeller2010} consider the case where $\wb_0$ is initialized in a quadratic basin and then analyze the minimum time $t$ such that $\wb_t$ is outside that basin.

They show that this so-called \emph{first exit time} depends \emph{exponentially} on the height of the basin; however, this dependency is only \emph{polynomial} with the width of the basin. These theoretical results directly contradict with the wide minima phenomenon: even if the height of a basin is slightly larger, the exit-time from this basin will be dominated by its height, which implies that the process would stay longer in (or in other words, `prefer') deeper minima as opposed to wider minima. The reason why the exit-time is dominated by the height is due to the \emph{continuity} of the Brownian motion, which is in fact a direct consequence of the Gaussian noise assumption.

A final remark on the issues of this approach is the observation that landscape is flat at the bottom regardless of the batch size used in SGD \citep{sagun2017empirical}. In particular, the spectrum of the Hessian at a near critical point with close to zero loss value has many near zero eigenvalues. Therefore, local curvature measures that are used as a proxy for measuring the width of a basin can be misleading. Such measures usually correlate with the magnitudes of large eigenvalues of the Hessian which are few \citep{keskar2016large,jastrzkebski2017three}. Besides, during the dynamics of SGD it has been observed that the algorithm does not cross barriers except perhaps at the very initial phase \citep{xing2018walk,Baity18}. Such dependence of width on an essentially-flat landscape combined with the lack of explicit barrier crossing during the SGD descent forces us to rethink the analysis of basin hopping under a noisy dynamics.

\subsection{Proposed framework}

In this study, we aim at addressing these contradictions and come up with an arguably better-suited hypothesis for the stochastic gradient noise that has more pertinent theoretical implications for the phenomena associated with SGD. In particular, we go back to \eqref{eqn:stoch_grad} and \eqref{eqn:noise_gauss} and reconsider the application of CLT. This \emph{classical} CLT assumes that $U_k$ is a sum of many independent and identically distributed (i.i.d.)\ random vectors, whose covariance matrix exists and is invertible,
and then it states that the law of $U_k$ converges to a Gaussian distribution, which then paves the way for \eqref{eqn:sgd_gauss}. 
Even though the finite-variance assumption seems natural and intuitive at the first sight, it turns out that in many domains, such as turbulent motions (\cite{weeks1995observation}), oceanic fluid flows (\cite{woyczynski2001levy}), finance (\cite{mandelbrot2013fractals}), biological evolution (\cite{jourdain2012levy}), audio signals (\cite{liutkus2015generalized,csimcsekli2015alpha,leglaive2017alpha,csimcsekli2018alpha}), brain signals (\cite{jas2017learning}), the assumption might fail to hold (see \citep{duan} for more examples). In such cases, the classical CLT along with the Gaussian approximation will no longer hold. While this might seem daunting, fortunately, one can prove a \emph{generalized} CLT and show that the law of the sum of these i.i.d.\ variables with infinite variance still converges to a family of \emph{heavy-tailed} distributions that is called the $\alpha$-stable distribution \citep{paul1937theorie}. As we will detail in Section~\ref{sec:levy_sde}, these distributions are parametrized by their \emph{tail-index} $\alpha \in (0,2]$ and they coincide with the Gaussian distribution when $\alpha =2$. 

In this study, we relax the finite-variance assumption on the stochastic gradient noise and by invoking the generalized CLT, we assume that $U_k$ follows an $\alpha$-stable distribution, as hinted in Figure~\ref{fig:hist_sas}. By following a similar rationale to \eqref{eqn:sgd_gauss} and \eqref{eqn:sgd_langevin}, we reformulate SGD with this new assumption and consider its continuous-time limit for small step-sizes. Since the noise might not be Gaussian anymore (i.e.\ when $\alpha \neq 2$), the use of the Brownian motion would not be appropriate in this case and we need to replace it with the $\alpha$-stable L\'{e}vy motion, whose increments have an $\alpha$-stable distribution (\cite{yanovsky2000levy}). Due to the heavy-tailed nature of $\alpha$-stable distribution, the L\'{e}vy motion might incur large discontinuous jumps and therefore exhibits a fundamentally different behavior than the Brownian motion, whose paths are on the contrary almost surely continuous. As we will describe in detail in Section~\ref{sec:levy_sde}, the discontinuities also reflect in the metastability properties of L\'{e}vy-driven SDEs, which indicate that, as soon as $\alpha <2$, the first exit time from a basin does \emph{not} depend on its height; on the contrary, it directly depends on its width and the tail-index $\alpha$. Informally, this implies that the process will \emph{escape} from narrow minima -- no matter how deep they are -- and stay longer in wide minima. Besides, as $\alpha$ gets smaller, the probability for the dynamic to jump into a wide basin will increase.
Therefore, if the $\alpha$-stable assumption on the stochastic gradient noise holds, then the existing metastability results automatically provide strong theoretical insights for illuminating the behavior of SGD.

\subsection{Contributions}

The main contributions of this paper are twofold: (i) we perform an extensive empirical analysis of the tail-index of the stochastic gradient noise in deep neural networks and (ii) based on these empirical results, we bring an alternative perspective to the existing approaches for analyzing SGD and shed more light on the folklore that SGD prefers wide minima by establishing a bridge between SGD and the related theoretical results from statistical physics and stochastic analysis.

We conduct experiments on the most common deep learning architectures. In particular, we investigate the tail behavior under fully-connected and convolutional models using negative log likelihood (NLL) and linear hinge loss functions on MNIST, CIFAR10, and CIFAR100 datasets. 
For each configuration, we scale the size of the network and batch size used in SGD and monitor the effect of each of these settings on the tail index $\alpha$. 

Our experiments reveal several remarkable results:
\begin{itemize}[noitemsep,topsep=1pt,leftmargin=*,align=left]
\item In all our configurations, the stochastic gradient noise turns out to be highly non-Gaussian and possesses a heavy-tailed behavior.
\item Increasing the size of the minibatch has a very little impact on the tail-index, and as opposed to the common belief that larger minibatches result in Gaussian gradient noise, the noise is still far from being Gaussian.
\item There is a strong interaction between the network architecture, network size, dataset, and the tail-index, which ultimately determine the dynamics of SGD on the training surface. This observation supports the view that, the geometry of the problem and the dynamics induced by the algorithm cannot be separated from each other. 
\item In almost all configurations, we observe two distinct phases of SGD throughout iterations. During the first phase, the tail-index rapidly decreases and SGD possesses a clear jump when the tail-index is at its lowest value and causes a sudden jump in the accuracy. This behavior strengthens the view that SGD crosses barriers at the very initial phase.
\end{itemize}

The paper is organized as follows. In Section~\ref{sec:levy_sde} we provide the technical background required for the $\alpha$-stable distributions and SDEs driven by $\alpha$-stable L\'{e}vy motions. We also formalize the framework in which we analyze SGD by using such SDEs as a proxy. We then describe in Section~\ref{sec:metastability} the metastability and first exit time properties of such SDEs and their discretizations, and discuss their connection with the wide minima phenomenon. In Section~\ref{sec:convergence}, we provide formal theoretical analysis for the convergence behavior of SGD to a local optimum under heavy-tailed gradient noise and discuss the implications of this result. In Section~\ref{sec:exps} we describe our experimental methodology and in Section~\ref{sec:experimentResult} we provide our empirical results which validate our theory. 
Our approach also opens up several interesting future directions and open questions, as we discuss in Section~\ref{sec:conc}.

We note that an initial version of this paper was presented at the 36th International Conference on Machine Learning \citep{simsekli_tail_ICML2019}. In this article, we have significantly extended our conference proceeding. These extensions can be summarized as follows:
\begin{itemize}[noitemsep,topsep=1pt,leftmargin=*,align=left]
    \item We have updated Section~\ref{sec:metastability} and added more detailed explanations for clarity.
    \begin{itemize}
        \item In addition to the metastability results, we have added a discussion of the first exit time properties of the L\'{e}vy-driven SDEs, which form the basis of the metastability results (Theorem~\ref{thm-exit-sde}). Furthermore, we have summarized our recent theoretical findings \citep{nguyen2019first} on the first exit time properties of discretized processes and translated them to the context of this article (Section~\ref{sec:discretization}). 
    \item We have added a summary of the first exit time behavior of the L\'{e}vy-driven systems in $\mathbb{R}^d$ (end of Section~\ref{sec:firstexitd1}).
    \end{itemize}
    \item In Section~\ref{sec:convergence}, under certain assumptions, we have proved a new local convergence result for SGD with an explicit rate and made a connection between the convergence rate of SGD and the tail-index of the gradient noise. At the end of Section~\ref{sec:convergence}, we have also added a short discussion about the global convergence properties of the discretized process by summarizing the results that we proved in \citep{nguyen2019non}.
    \item In addition to tail-index estimation, we have also considered a statistical test for determining the stability of a process \citep{brcich2005stability} (Section~\ref{sec:stableTest}).
    \item Finally, we have added several new experimental results (Section~\ref{sec:experimentResult}). In particular, we have presented new results on stability tests (Section~\ref{sec:stability_test_res}), finer-grained layer-wise tail-index estimation (Sections~\ref{sec:exps_size} and \ref{subsec:exp_iter}), and an investigation of the relation between the tail-index and the generalization properties of the network (\ref{sec:gen}).
\end{itemize}

\section{Stable distributions and SGD as a L\'{e}vy-Driven SDEs}
\label{sec:levy_sde}

The CLT states that the sum of i.i.d. random variables with a finite second moment converges to a normal distribution if the number of summands grow. However, if the variables have heavy-tails, the second moment may not exist. For instance, if their density $p(x)$ has a power-law tail decreasing as $1/|x|^{\alpha+1}$ where $0 < \alpha < 2$; only $r$-th moment exists with $r<\alpha$. In this case, generalized central limit theorem (GCLT) says that the sum of such variables will converge to a distribution called the  \emph{$\alpha$-stable} distribution instead as the number of summands grows (see e.g. \citep{fischer2010history}. In this work, we focus on the centered \emph{symmetric $\alpha$-stable} ($\sas$) distribution, which is a special case of $\alpha$-stable distributions that are symmetric around the origin.

We can view the $\sas$ distribution as a heavy-tailed generalization of a centered Gaussian distribution. The $\sas$ distributions are defined through their characteristic function via 
\begin{align}
X\sim \sas(\sigma) \iff \E[\exp(i \omega X)] = \exp(-|\sigma \omega|^\alpha).    
\end{align}
 Even though their probability density function does not admit a closed-form formula in general except in special cases, their density decays with a power law tail like $1/|x|^{\alpha+1}$ where $\alpha \in (0,2]$ is called the \emph{tail-index} which determines the behavior of the distribution: as $\alpha$ gets smaller; the distribution has a heavier tail. In fact, the parameter $\alpha$ also determines the moments: when $\alpha<2$, $\mathds{E}[|X|^r] < \infty$ if and only if $r<\alpha$; implying $X$ has infinite variance when $\alpha\neq 2$. The parameter $\sigma \in \mathds{R}_+$ is the \emph{scale} parameter and controls the spread of $X$ around $0$. We recover the Gaussian distribution ${\cal N}(0,2\sigma^2)$ as a special case when $\alpha=2$ and the Cauchy distribution when $\alpha =1$.

Following the above argument, a more general assumption on the stochastic gradient noise can be given by:
\begin{align}
[U_k(\wb)]_i \sim \sas_i(\sigma_i(\wb)), \quad \forall i =1,\dots,n \label{eqn:noise_levy} 
\end{align}
where $[v]_i$ denotes the $i$'th component of a vector $v$, and $\sas_i$ distributed with $\alpha_i(w)$. Clearly, this assumption is way too general to offer reasonable theoretical treatment. We will resort to several simplifications: (1) We assume that each coordinate of $U_k$ is $\sas$ distributed with the same $\sigma$ which depends on the state $\wb$. Here, this dependency is not crucial since we are mainly interested in the tail-index $\alpha$, which can be estimated \emph{independently} from the scale parameter. Therefore, we will simply denote $\sigma(\wb)$ as $\sigma$ for clarity. (2) We further assume that each coordinate of $U_k$ is $\sas$ distributed with the same $\alpha$ independent of the state $\wb$. We will demonstrate the state independence at later stages of SGD experimentally in Section~\ref{subsec:exp_iter}, however, imposing the coordinate dependence is a much harder challenge which will be addressed in the section devoted for open problems (Section~\ref{sec:conc}).

\begin{figure}[t]
    \centering
    \includegraphics[width=0.49\columnwidth]{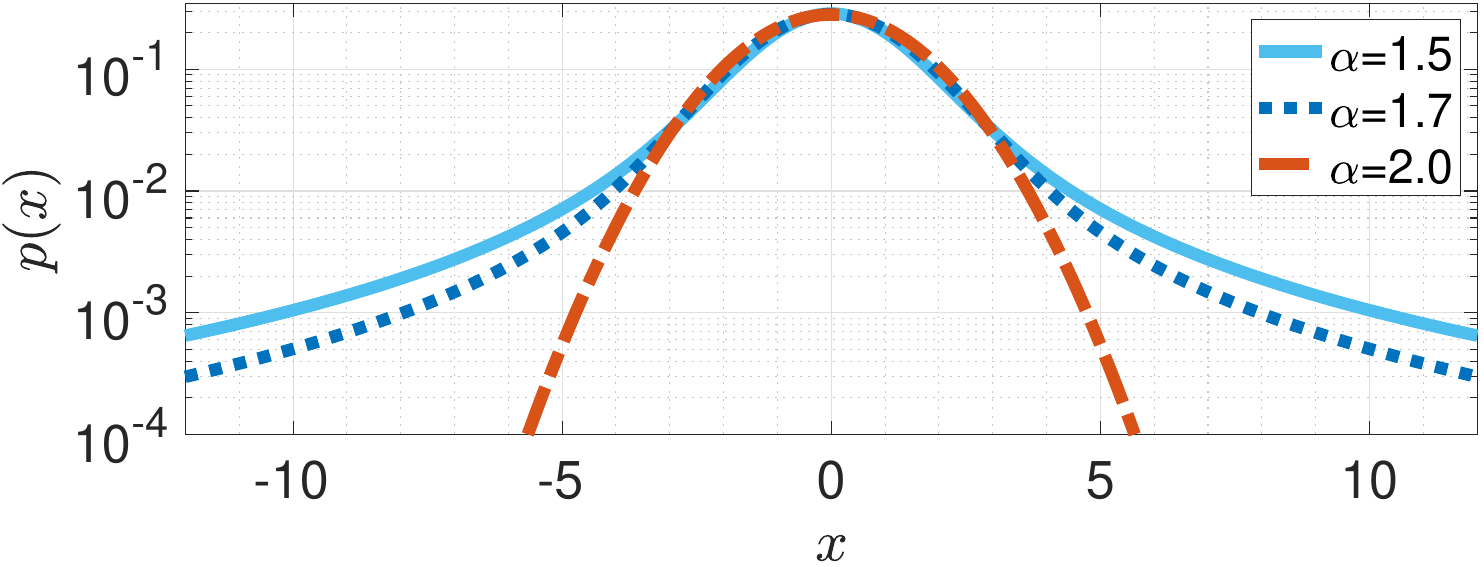} \hfill
    \includegraphics[width=0.495\columnwidth]{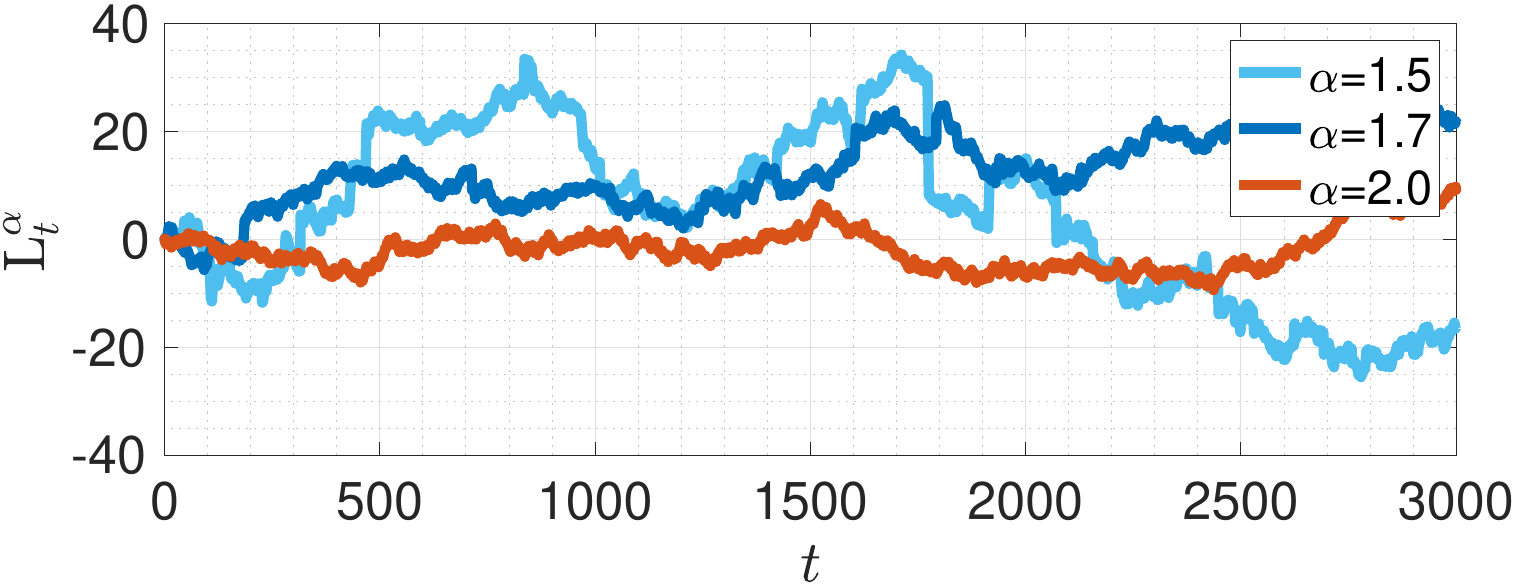}
    \caption{Left: $\sas$ densities, right:  $\mathrm{L}_t^\alpha$ for $d=1$. For $\alpha<2$, $\sas$ becomes heavier-tailed and $\mathrm{L}_t^\alpha$ incurs jumps.}
    \vspace{-10pt}
    \label{fig:sas_lm}
\end{figure}

By using the assumption \eqref{eqn:noise_levy}, we can rewrite the SGD recursion as follows \citep{csimcsekli2017fractional,nguyen2019non}:
\begin{align}
\wb^{k+1} = \wb^{k} - \eta \nabla f(\wb^k) + \eta^{1/\alpha} \Bigl(\eta^{\frac{\alpha-1}{\alpha} } \sigma\Bigr) S_k, \label{eqn:sgd_alpha}
\end{align}
where $S_k \in \mathbb{R}^d$ is a random vector such that $[S_k]_i \sim \sas(1)$. If the step-size $\eta$ is small enough, then we can consider the continuous-time limit of this discrete-time process, which is expressed in the following SDE driven by an $\alpha$-stable L\'{e}vy process:
\begin{align}
\rmd \wb_t = - \nabla f(\wb_t) \rmd t + \eta^{(\alpha-1)/\alpha} \sigma \> \rmd \Lm_t, \label{eqn:sgd_levy}
\end{align}
where $\Lm_t$ denotes the $d$-dimensional $\alpha$-stable L\'{e}vy motion with \emph{independent components}. In other words, each component of $\Lm_t$ is an independent $\alpha$-stable L\'{e}vy motion in $\mathbb{R}$. For the scalar case it is defined as follows for $\alpha \in (0,2]$ (\cite{duan}):

\begin{enumerate}[label=(\roman*),itemsep=0pt,topsep=0pt,leftmargin=*,align=left]
\item $\Lm_0 = 0$ almost surely.
\item For $t_0<t_1 < \cdots < t_N$, the increments $ (\Lm_{t_{i}} - \Lm_{t_{i-1}} )$ are independent ($i = 1,\dots, N$). %
\item The difference $(\Lm_t - \Lm_s)$ and $\Lm_{t-s}$ have the same distribution: $\sas((t-s)^{1/\alpha})$ for $s<t$. 
\item $\Lm_t$ is continuous in probability (i.e.\ it has \emph{stochastically continuous} sample paths): for all $\delta >0$ and $s\geq 0$, $p(|\Lm_t - \Lm_s| > \delta) \rightarrow 0$ as $t \rightarrow s$.
\end{enumerate}
{ It is easy to check that the noise term in \eqref{eqn:sgd_alpha} is obtained by integrating $\Lm_t$ from $k\eta$ to $(k+1)\eta$. When $\alpha = 2$, $\Lm_t$ coincides with a scaled version of Brownian motion, $\sqrt{2} \Bm_t$. $\sas$ and $\Lm_t$ are illustrated in Figure~\ref{fig:sas_lm}.}

The SDE in \eqref{eqn:sgd_levy} exhibits a fundamentally different behavior than the one in \eqref{eqn:sgd_langevin} does. This is mostly due to the stochastic continuity property of $\Lm_t$, which enables $\Lm_t$ to have a countable number of discontinuities, which are sometimes called `jumps'.
 In the rest of this section, we will recall important theoretical results about this SDE and discuss their implications on SGD.
\vspace{0.1cm}
\section{First Exit Time and Metastability Analysis}
\label{sec:metastability}

We start by reviewing known metastability properties of the $\alpha$-stable L\'evy process \eqref{eqn:sgd_levy} from the literature. We will also focus on the \emph{first exit time} which is, roughly speaking, the average time it takes for the process to exit a neighborhood of a local minima (a quantity we define formally later in \eqref{eqn:fet_cont-dimone}). Then, we will summarize the theoretical results proven in our recent work \citep{nguyen2019non} on the metastability properties of the discrete-time processes obtained by an Euler discretization of such processes.

\subsection{The continuous-time process}
\label{sec:firstexitd1}
For clarity of the presentation and notational simplicity we focus on the scalar case and consider the SDE \eqref{eqn:sgd_levy} in $\rset$ (i.e.\ $d=1$). Multidimensional generalizations of the metastability results presented in this paper can be found in \citep{imkeller2010first} and will be summarized at the end of this section. We rewrite \eqref{eqn:sgd_levy} as follows:
\beq \label{eq-levy-sde} \rmd  \wb_t^\varepsilon = - f'(\wb_t^\varepsilon) \rmd t + \varepsilon \rmd \Lm_t  
\eeq
for $t\geq 0$, started from the initial point $w_0\in\mathbb{R}$, where $\Lm_t$ is the $\alpha$-stable L\'evy process, $\varepsilon \geq 0$ is the noise level
and $f$ is a non-convex objective with $r \geq 2$ local minima. We denote the derivative of $f$ by $f'$. When $\varepsilon=0$, we recover the gradient descent dynamics in continuous time: $\rmd  \wb_t^0 = -f'(\wb_t^0) \rmd t$,
where the local minima are the stable points of this differential equation. However, as soon as $\varepsilon >0$, these states become `metastable', meaning that there is a positive probability for $\wb_t^\varepsilon$ to transition from one basin to another. However, the time required for transitioning to another basin strongly depends on the characteristics of the injected noise. 
The two most important cases are $\alpha =2$ and $\alpha < 2$. When $\alpha =2$, (i.e.\ the Gaussianity assumption) the process $(\wb^\varepsilon_t)_{t \geq 0}$ is continuous, which requires it to `climb' the basin all the way up, in order to be able to transition to another basin. This fact makes the transition-time depend on the height of the basin. On the contrary, when $\alpha <2$, the process can incur discontinuities and does not need to cross the boundaries of the basin in order to transition to another one, since it can directly jump. This property is called the `transition phenomenon' \citep{duan} and makes the transition-time mostly depend on the \emph{width} of the basin. In the rest of the section, we will formalize these explanations.

{ Gradient-like flows driven by Brownian motion and weak error for their discretization are well studied from a theoretical standpoint (see e.g. \citep{li2017stochastic,mertikopoulos2018convergence}), however their {L\'evy}-driven analogue \eqref{eq-levy-sde} and the discrete-time versions \citep{burghoff2015spectral} are relatively less studied.}
Under some assumptions on the objective $f$, it is known that the process \eqref{eq-levy-sde} admits a stationary density \citep{Samorodnitsky2003}. For a general $f$, an explicit formula for the equilibrium distribution is not known, however when the noise level $\varepsilon$ is small enough, finer characterizations of the structure of the equilibrium density in dimension one is known. We next summarize known results in this area, which show that L\'{e}vy-driven dynamics spend more time in `wide valleys' in the sense of \citep{entropy-sgd} when $\varepsilon$ goes to zero.

\begin{wrapfigure}{r}{0.4\columnwidth}
    \centering
    \label{fig:double_well} 
    \includegraphics[width=0.35\columnwidth]{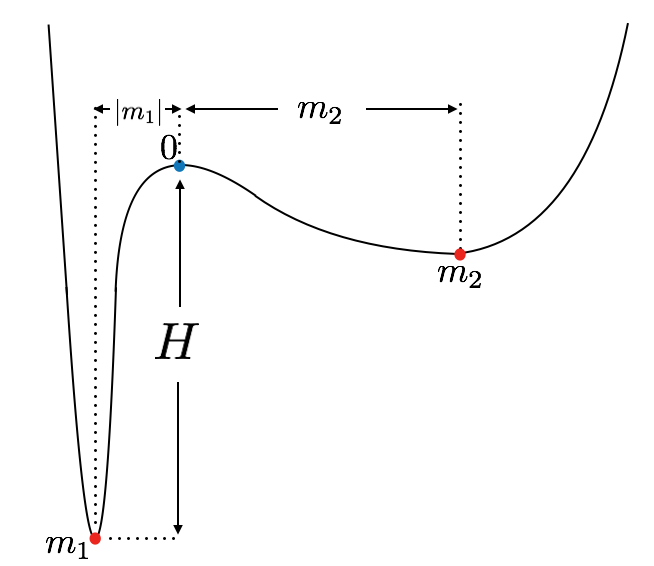}
    \caption{ An objective with two local minima $m_1, m_2$ seperated by a local maxima at $s_1 = 0$. 
    }
\end{wrapfigure}

Assume that $f$ is smooth with $r$ local minima $\{m_i\}_{i=1}^r$ separated by $r-1$ local maxima $\{s_i\}_{i=1}^{r-1}$, i.e. 
\beqs -\infty := s_0 < m_1 < s_1  < \dots <s_{r-1} < m_r < s_r := \infty. \eeqs
Furthermore, assume that the local minima and maxima are not degenerate, i.e. $f''(m_i)>0$ and $f''(s_i)<0$ for every $i$. We also assume the objective gradient has a growth condition $f'(w) >|w|^{1+c}$ for some constant $c>0$ and when $|w|$ is large enough. Each local minima $m_i$ lies in the (interval) valley $S_i = (s_{i-1},s_i)$ of (width) length $L_i = |s_i-s_{i-1}|$. Consider also a $\delta$-neighborhood $B_i := \{ |x - m_i|\leq \delta \}$ around the local minimum with $\delta>0$ small enough so that the neighborhood is contained in the valley $S_i$ for every $i$. We are interested in the first exit time from $B_i$ starting from a point $w_0\in B_i$ and the transition time $T_{w_0}^i(\varepsilon):= \inf \{ t\geq 0 : \wb_t^\varepsilon \in \cup_{j\neq i} B_j \}$ to a neighborhood of another local minimum, we will remove the dependency to $w_0$ of the transition time in our discussions as it is clear from the context. The following result shows that the transition times are asymptotically exponentially distributed in the limit of small noise and scales like $1/{\varepsilon^\alpha}$ with $\varepsilon$.
 \begin{theorem}[\cite{pavlyukevich2007cooling}]\label{thm-levy-exit} For an initial point $w_0\in B_i$, in the limit $\varepsilon\to0$, the following statements hold regarding the transition time:
   \beqs 
         \mathbb{P}_{w_0}(T^i(\varepsilon) \in B_j) &\to& q_{ij} q_i^{-1} \quad \mbox{if} \quad i\neq j,  \\
         \mathbb{P}_{w_0}(\varepsilon^\alpha T^i(\varepsilon) \geq u ) &\leq& e^{-q_i u} \quad \mbox{for any} \quad u\geq 0.
  \eeqs
 where
  \beq    
         q_{ij} &=& \frac{1}{\alpha} \left| \frac{1}{|s_{j-1} - m_i |^\alpha} - \frac{1}{|s_{j} - m_i |^\alpha} \right|, \\
         q_{i} &=& \sum_{j\neq i}q_{ij}.
     \eeq
     \vspace{-20pt}
 \end{theorem}
 If the SDE \eqref{eq-levy-sde} would be driven by the Brownian motion instead, then an analogous theorem to Theorem \ref{thm-levy-exit} holds saying that the transition times are still exponentially distributed but the scaling $\varepsilon^\alpha$ needs to be replaced by $e^{2H/\varepsilon^2}$ where $H$ is the maximal depth of the basins to be traversed between the two local minima \citep{Day83,bovier2005metastability}. This means that in the small noise limit, Brownian-motion driven gradient descent dynamics need exponential time to transit to another minimum whereas Levy-driven gradient descent dynamics need only polynomial time. We also note from Theorem \ref{thm-levy-exit} that the mean transition time between valleys for L\'evy SDE does not depend on the depth $H$ of the valleys they reside in which is an advantage over Brownian motion driven SDE in the existence of deep valleys. Informally, this difference is due to the fact that Brownian motion driven SDE has to typically climb up a valley to exit it, whereas L\'{e}vy-driven SDE could jump out.

The following theorem says that as $\varepsilon \to 0$, up to a normalization in time, the process $w_t^\varepsilon$ behaves like a finite state-space Markov process that has support over the set of local minima $\{m_i\}_{i=1}^r$ admitting a stationary density $\pi = (\pi_i)_{i=1}^r$ with an infinitesimal generator $Q$. The process jumps between the valleys $S_i$, spending time proportional to probability $p_i$ amount of time in each valley in the equilibrium where the probabilities $\pi = (\pi_i)_{i=1}^r$ are given by the solution to the linear system $Q\pi = 0$.   
\begin{theorem}[{\cite{pavlyukevich2007cooling}}]\label{thm-levy-exit-2} Let $w_0 \in S_i$, for some $ 1\leq i \leq r$. For $t\geq 0$, 
   $\wb_{t\varepsilon^{-\alpha}}^\varepsilon \to Y_{m}(t)$, as $\varepsilon\to 0$,
in the sense of finite-dimensional distributions, where $Y = (Y_{m}(t))_{t\geq 0}$ is a continuous-time Markov chain on a state space $\{m_1,m_2,\dots,m_r\}$ with the infinitesimal generator $Q = (q_{ij})_{i,j=1}^r$ with 
     \begin{align}
             q_{ij} &= \frac{1}{\alpha} \left| \frac{1}{|s_{j-1} - m_i |^\alpha} - \frac{1}{|s_{j} - m_i |^\alpha} \right|, \\
         q_{ii} &=-\sum _{j\neq i} q_{ij}. 
     \end{align}    
This process admits a density $\pi$ satisfying $Q^T\pi = 0$.    
\end{theorem}

A consequence of this theorem is that equilibrium probabilities $\pi_i$ are typically larger for ``wide valleys". To see this consider the special case illustrated in Figure \ref{fig:double_well} with $r=2$ local minima $m_1 < s_1 = 0 < m_2$ separated by a local maximum at $s_1 = 0$. For this example, $m_2 > |m_1|$, and the second local minimum lies in a wider valley. A simple computation reveals 
  $$ \pi_1 = \frac{|m_1|^\alpha}{|m_1|^\alpha + m_2^\alpha}, \quad \pi_2 = \frac{|m_2|^\alpha}{|m_1|^\alpha + |m_2|^\alpha}. $$
We see that $\pi_2 > \pi_1$, that is in the equilibrium the process spends more time on the wider valley. In particular,  the ratio $\frac{\pi_2}{\pi_1} = \left(\frac{m_2}{|m_1|}\right)^\alpha$
grows with an exponent $\alpha$ when the ratio $\frac{m_2}{|m_1|}$ of the width of the valleys grows. 
Consequently, if the gradient noise is indeed $\alpha$-stable distributed, these results directly provide theoretical evidence for the wide-minima behavior of SGD assuming the loss landscape is not degenerate. 

In addition to the transition time between the basins of attraction of two local minima, understanding how long it takes for the continuous-time process $w_t$ given by \eqref{eqn:sgd_levy} to exit a neighborhood of a local minimum $\bar{w}$ (given that it is started in that neighborhood) is also relevant. We formally define the \emph{first exit time} of the stochastic process \eqref{eqn:sgd_levy} as follows: 
\begin{align}
\label{eqn:fet_cont-dimone} \tau_{a}(\varepsilon)&\triangleq\inf\{t\geq 0: | w_t -\bar{w}|\not\in[0,a]\}.
\end{align} 
The following result characterizes the first exit time in dimension one. \begin{theorem}[\cite{imkeller2006first}]\label{thm-exit-sde}
Consider the SDE \eqref{eqn:sgd_levy} in dimension $d=1$ and assume that it has a unique strong solution. Assume further that the objective $f$ has a global minimum at zero, satisfying the conditions $f'(x)x\geq0$ {for every $x\in\mathbb{R}$}, $f(0)=0$, $f'(x) = 0$ if and only if $x=0$, and $f''(0) >0$. Then, there exist positive constants $\varepsilon_0$, $\gamma$, $\delta$, and $C>0$ such that for $0 < \varepsilon \leq \varepsilon_0$, the following holds:
\begin{align}
\mathrm{e}^{-u \varepsilon^\alpha \frac{\theta}{\alpha} (1+ C \varepsilon^\delta) } (1-  C \varepsilon^\delta)  \leq \mathbb{P}( \tau_{a}(\varepsilon) >u ) \leq \mathrm{e}^{-u \varepsilon^\alpha \frac{\theta}{\alpha} (1 - C \varepsilon^\delta) } (1 +  C \varepsilon^\delta)
\end{align}
uniformly for all {initialization $w_0$}$ \in [-a + \varepsilon^\gamma , a - \varepsilon^\gamma] $ and $u \geq 0 $, where %
$\theta = \frac2{a^\alpha}$. Consequently, 
\begin{align}\label{eq-tau-eps}
\mathbb{E}[\tau_{a}(\varepsilon)] = \frac{\alpha}{2}\frac{a^\alpha}{\varepsilon^\alpha} (1+ \mathcal{O}(\varepsilon^\delta)), \quad \text{ uniformly for all } {w_0 \in [-a + \varepsilon^\gamma , a - \varepsilon^\gamma]}. 
\end{align}
\end{theorem}

\noindent \textbf{The case of $\mathbb{R}^d$.} The exit behavior of the SDE \eqref{eqn:sgd_levy} from an arbitrary domain in $\mathbb{R}^d$ has also been studied in the literature. \cite{imkeller2010first} generalizes Theorem \ref{thm-exit-sde} from dimension $d=1$ to arbitrary dimensions and showed that in the small  noise limit the exit time from a domain is exponentially distributed with a parameter that depends on the tail-index $\alpha$. In case the components of the L{\'e}vy motion in \eqref{eqn:sgd_levy} is replaced by a process that consists of the sum of finitely many one-dimensional L{\'e}vy processes with different tail-indices $\alpha_i$, it is also shown that the first exit time from a domain is determined by the smallest $\alpha_i$ when the noise level $\varepsilon$ is small enough.

\subsection{Relating the discretization to the continuous-time process}
\label{sec:discretization}
While the metastability and first exit time results of L\'{e}vy-driven SDEs can be used as a proxy for analyzing SGD, approximating SGD as a continuous-time approach might not be accurate for any step-size $\eta$, and some theoretical concerns have already been raised for the validity of such approximations (\cite{yaida2018fluctuationdissipation}). Intuitively, one can expect that the metastable behavior of SGD would be similar to the behavior of its continuous-time limit only when the discretization step-size is small enough. Even though some theoretical results have been recently established for the discretizations of SDEs driven by Brownian motion (\cite{tzen2018local}), it is not clear how the discretized L\'{e}vy SDEs behave in terms of metastability. 

In this section, we summarize the theoretical results that we proved in our recent work \citep{nguyen2019first}. In particular, we will now present explicit conditions for the step-size such that the metastability behavior of the discrete-time system \eqref{eqn-sgd-discrete-model} is guaranteed to be close to its continuous-time limit \eqref{eqn-general-sde}. More precisely, we consider the following stochastic differential equation with both a Brownian term and a L\'evy term, and its Euler discretization as follows (\cite{duan}):
    \begin{align}
    \label{eqn-general-sde} 
    \mathrm{d}\wb_t &=-\nabla f(\wb_{t})\mathrm{d}t + \varepsilon\sigma \mathrm{d} \Bm_t + \varepsilon\mathrm{d} \Lm_t \\
    \label{eqn-sgd-discrete-model}
\wb^{k+1} &= \wb^k-\eta \nabla f(\wb^k) + \varepsilon\sigma\eta^{1/2}Z_{k+1} + \varepsilon\eta^{1/\alpha}S_{k+1},
    \end{align}
with independent and identically distributed (i.i.d.) variables $Z_k \sim \mathcal{N}(0, I)$ where $I$ is the identity matrix, the components of $S_k$ are i.i.d with $\mathcal{S}\alpha\mathcal{S}(1)$ distribution, and $\varepsilon$ is the amplitude of the noise. This dynamics includes \eqref{eqn:sgd_langevin} and  \eqref{eqn:sgd_levy} as special cases. 
Here, $\sigma$ is chosen as a scalar for convenience; however, we believe that this analysis can be extended to the case where $\sigma$ is a function of $\wb_t$.

Understanding the metastability behavior of SGD modeled by these dynamics requires understanding the first exit times for the continuous-time process $\wb_t$ given by \eqref{eqn-general-sde} and its discretization $\wb^k$ \eqref{eqn-sgd-discrete-model}. 
For this purpose, for any given local minimum $\bar{\wb}$ of $f$ and $a>0$, we define the following set 
\begin{align}
A\triangleq\Big\{(\wb^1,\ldots,\wb^K)\in\mathbb{R}^d\times\ldots\times\mathbb{R}^d:\max_{k\leq K}\Vert \wb^k-\bar{\wb}\Vert\leq a\Big\},
\label{eqn:set_A}
\end{align}
which is the set of $K$ points in $\mathbb{R}^d$, each at a distance of at most $a$ from the local minimum $\bar{\wb}$. 
Similar to \eqref{eqn:fet_cont-dimone}, we will study the first exit times defined by
\begin{align}
\label{eqn:fet_cont} \tau_{\xi,a}(\varepsilon)&\triangleq\inf\{t\geq 0:\Vert \wb_t-\bar{\wb}\Vert\not\in[0,a+{\xi}]\}, \\
\label{eqn:fet_dis} \bar{\tau}_{\xi,a} (\varepsilon)&\triangleq\inf\{k\in\mathbb{N} \> : \> \Vert \wb^k-\bar{\wb}\Vert\not\in[0,a+{\xi}]\}.
\end{align}
Note that in the special case $\xi=0$, we recover ${\tau}_{0,a}(\varepsilon)=\tau_a(\varepsilon)$ introduced previously in \eqref{eqn:fet_cont-dimone}. 

Our result (Theorem~\ref{thm:ultimateTheorem}) shows that with sufficiently small discretization step $\eta$, the probability to exit a given neighborhood of the local optimum at a fixed time $t$ of the discretization process approximates that of the continuous process. This result also provides an explicit condition for the step-size, which explains certain impacts of the other parameters of the problem, such as dimension $d$, noise amplitude $\varepsilon$, variance of Gaussian noise $\sigma$, towards the similarity of the discretization and continuous processes.

Let us now state the assumptions which will imply our result.
\begin{assumption}
\label{assump:unq} 
The SDE \eqref{eqn-general-sde} admits a unique strong solution. 
\end{assumption}
\begin{assumption} 
\label{assump:novikov}
Consider the process
$\mathrm{d}\hat{\wb}_t=g(\hat{\wb})\mathrm{d}t + \varepsilon\sigma \mathrm{d} \Bm_t + \varepsilon\mathrm{d} \Lm_t$,
where $\hat{\wb} \equiv \{\hat{\wb}_t\}_{t\geq 0}$ denotes the whole process and the drift $g$ is defined as follows\footnote{It is easy to verify that $\hat{\wb}_{k\eta} = \wb^k$ for all $k \in \mathbb{N}_+$ \citep{raginsky17a}.}:
\begin{align*}
g (\hat{\wb}) &\triangleq -\sum\limits_{k=0}^{\infty}\nabla f(\hat{\wb}_{k\eta})\mathbb{I}_{[k\eta,(k+1)\eta)}(t). 
\end{align*}
Here, $\mathbb{I}$ denotes the indicator function, i.e.\ $\mathbb{I}_{S}(x) = 1$ if $x \in S$ and $\mathbb{I}_{S}(x) = 0$ if $x \notin S$.
Then, the process $\phi_t \triangleq -\frac{g({\wb}) + \nabla f(\wb_t)}{\varepsilon \sigma}$ satisfies:
$\mathbb E \exp\left(\frac{1}{2}\int_0^T \phi^2_t \mathrm{d}t\right)<\infty.$ 
\end{assumption}
\begin{assumption}\label{assump:holderCondition}
The gradient of $ f$ is $\gamma$-H\"{o}lder continuous: There exists a constant $M>0$ such that 
\begin{align*}
\|\nabla f(x) - \nabla f(y)\| \leq M \|x-y \|^\gamma, \qquad \forall x,y\in\mathbb{R}^d.
\end{align*}
\end{assumption}
\begin{assumption}\label{assump:gradientCondition}
The gradient of $f$ satisfies the following assumption: 
$\|\nabla f(0) \| \leq B$. 
\end{assumption}
\begin{assumption}\label{assump:dissipative}
For some $m>0$ and $b\geq 0$, $f$ is $(m,b,\gamma)$-dissipative:
\begin{align*}
\langle x,\nabla f(x)\rangle\geq m\Vert x\Vert^{1+\gamma}-b, \quad \forall x\in\mathbb{R}^d.
\end{align*}
\end{assumption}

We note that, \textbf{A}\ref{assump:unq} has been a common assumption in stochastic analysis, e.g.\ \citep{imkeller2006first,imkeller2010first,liang2018gradient}
and \textbf{A}\ref{assump:unq} and \textbf{A}\ref{assump:novikov} directly hold for bounded gradients.
On the other hand, the assumptions \textbf{A}\ref{assump:holderCondition}-\textbf{A}\ref{assump:dissipative} are standard conditions, which are often considered in non-convex optimization algorithms that are based on discretization of diffusions \citep{raginsky17a,xu2018global,erdogdu2018global,gao18,gao-sghmc}.

The next assumption identifies an explicit condition for the step-size, which is required to make sure that the discrete process well-approximates the continuous one.
\begin{assumption}
\label{assump:stepsize}
For a given $\delta >0$, $t = K\eta$, and for some $C>0$, the step-size satisfies the following condition:
\begin{align*}
0<\eta\leq\min\Big\{1,\frac{m}{M^2}, \Big(\frac{\delta^2}{2K_1t^2}\Big)^{\frac{1}{\gamma^2+2\gamma-1}}, \Big(\frac{\delta^2}{2K_2t^2}\Big)^{\frac{1}{2\gamma}}, \Big(\frac{\delta^2}{2K_3t^2}\Big)^{\frac{\alpha}{2\gamma}}, \Big(\frac{\delta^2}{2K_4t^2}\Big)^{\frac{1}{\gamma}}\Big\},
\end{align*}
where $\varepsilon$ is as in \eqref{eqn-sgd-discrete-model}, the constants $m,M,b$ are defined by \textbf{A}\ref{assump:holderCondition}-- \textbf{A}\ref{assump:dissipative} and
\begin{align*}
K_1 = \mathcal{O}(d\varepsilon^{2\gamma^2-2}),\, \>\> K_2 = \mathcal{O}(\varepsilon^{-2}),\, \>\> K_3 = \mathcal{O}(d^{2\gamma}\varepsilon^{2\gamma-2}),\, \>\> K_4 = \mathcal{O}(d^{2\gamma}\varepsilon^{2\gamma-2}).
\end{align*}
\end{assumption}

More explicit forms of the constants are provided in \citep{nguyen2019first}. We then have the following theorem, associating the first exit times of the continuous and the discretized processes.
\begin{theorem}[\cite{nguyen2019first}]\label{thm:ultimateTheorem}
Under assumptions \textbf{A}\ref{assump:unq}- \textbf{A}\ref{assump:stepsize}, the following inequality holds:
\begin{align*}
\mathbb{P}[\tau_{-\xi,a}(\varepsilon)>K\eta]-C_{K,\eta,\varepsilon,d,\xi}-\delta\leq\mathbb{P}[\bar{\tau}_{0,a}(\varepsilon)>K]\leq\mathbb{P}[\tau_{\xi,a}(\varepsilon)>K\eta]+C_{K,\eta,\varepsilon,d,\xi}+\delta,
\end{align*}
where,
\begin{align*}
C_{K,\eta,\varepsilon,d,\xi}\triangleq& \frac{C_1(K\eta(d\varepsilon+1)+1)^\gamma e^{M\eta}M\eta}{\xi} + 1-\Big(1-Cde^{-\xi^2e^{-2M\eta}(\varepsilon\sigma)^{-2}/(16d\eta)}\Big)^K\\
&+ 1-\Big(1-C_\alpha d^{1+\alpha/2}\eta e^{\alpha M\eta}\varepsilon^{\alpha}\xi^{-\alpha}\Big)^K,
\end{align*}
for some constants $C_1,C_\alpha$ and $C$ that does not depend on $\eta$ or $\varepsilon$, 
$M$ is given by \textbf{A}\ref{assump:holderCondition} and $\varepsilon$ is as in \eqref{eqn-general-sde}--\eqref{eqn-sgd-discrete-model}. 
\end{theorem}

\noindent \textbf{Exit time versus problem parameters.} In Theorem~\ref{thm:ultimateTheorem}, if we let $\eta$ go to zero for any $\delta$ fixed, the constant $C_{K,\eta,\varepsilon,d,\xi}$ will also go to zero, and since $\delta$ can be chosen arbitrarily small, this implies that the probability of the first exit time for the discrete process and the continuous process will approach each other when the step-size gets smaller, as expected. If instead, we decrease $d$ or $\varepsilon$, the quantity $C_{K,\eta,\varepsilon,d,\xi}$ also decreases monotonically, but it does not go to zero due to the first term in the expression of $C_{K,\eta,\varepsilon,d,\xi}$. \vspace{5pt}

\noindent \textbf{Exit time versus width of local minima.} Popular activation functions used in deep learning such as ReLU functions are almost everywhere differentiable and therefore the cost function has a well-defined Hessian almost everywhere (see e.g. \citep{yang17relu}). The eigenvalues of the Hessian of the objective near local minima have also been studied in the literature (see e.g. \citep{sagun2016eigenvalues,papyan2018full}). If the Hessian around a local minimum is positive definite, the conditions for the multi-dimensional version of Theorem \ref{thm-exit-sde} in \citep{imkeller2010first}) are satisfied locally around a local minimum. For local minima lying in wider valleys, the parameter $a$ can be taken to be larger; in which case the expected exit time $\mathbb{E}\tau_{0,a}(\varepsilon)\sim \mathcal{O}(a^\alpha) $ will be larger by the formula \eqref{eq-tau-eps}. In other words, the SDE \eqref{eqn-general-sde}  spends more time to exit wider valleys. Theorem \ref{thm:ultimateTheorem} shows that SGD modeled by the discretization of this SDE will also inherit a similar behavior if the step-size satisfies the conditions we provide.

\section{Convergence Analysis}

\label{sec:convergence}

The convergence of SGD iterates to a local minimum in the context of deep learning has been studied in the literature (see e.g. \citep{reddi2016stochastic,wu2018wngrad}). However, these results apply only when the gradient noise has a finite second moment. A natural question that arises is how fast SGD iterates converge to a local minimum when the noise has tail index $\alpha\in(1,2]$. In this case, the second moment exists only when $\alpha=2$. For $\alpha<2$, the second moment does not exist but the $(1+\gamma)$-th moment exist for any $\gamma \in [0,\alpha-1)$. Recall that the stochastic gradient iterations with constant stepsize are of the form
\begin{equation}\label{iter-sgd}\wb^{k+1} = \wb^k - \eta  \nabla \tilde{f}_k(\wb^k)
\end{equation}
\noindent where $\eta$ is the stepsize, $ \nabla \tilde{f}_k(\wb^k)$ is the stochastic gradient and $\wb^0$ is the initialization. Consider the random gradient noise $U_k$ at step $k$.
Following the literature, we assume that $U_k$ is measurable with respect to an increasing family of Borel fields $\mathcal{F}_{k}$ defined on a probability space $\mathcal{P}$. We also assume that  %
the noise is unbiased and the stochastic gradients have a finite $(1+\gamma)$-th moment, i.e.
\begin{assumption}
$\mathbb{E} \left( U_k \big | \mathcal{F}_k \right) = 0 \quad \mbox{and} \quad  \mathbb{E} \left(\|\nabla \tilde{f}_k(\wb^k) \|^{1+\gamma} \big | \mathcal{F}_k \right) \leq \sigma_\gamma^{1+\gamma} $ 
for some $\sigma_\gamma >0$ and $\gamma < \alpha - 1$. 
\label{asmp:moment}
\end{assumption}
When $\gamma=1$ (which requires $\alpha=2$), we recover the standard setting studied in the literature (see e.g. \citep{reddi2016stochastic,polyak1992acceleration}). Here we consider $\gamma \in [0,\alpha-1)$ to account for potential heavy-tailedness in the gradient noise.

Under this assumption, we have the following convergence result for SGD. The proof is given in the appendix and is inspired by the proof technique of \citep{reddi2016stochastic} for the $\gamma=1$ and $\alpha=2$ case and extends it to the heavy-tailed case when $\gamma\in[0,\alpha-1)$ and $\alpha<2$.
\begin{theorem} 
\label{thm:local_conv}
Assume \cref{assump:holderCondition}-\cref{asmp:moment} hold with $\gamma \in [0,\alpha-1)$ and the objective is bounded below admitting a minimum $f_*$. Consider the SGD iterations \eqref{iter-sgd} with initial point $\wb^0$ and constant stepsize $\eta>0$. Then, we have
\begin{eqnarray}
\min_{0\leq k \leq K-1} \mathbb{E} \| \nabla f(\wb^k)\|^2 
 \leq  \frac{f(\wb^0) - f_*}{K\eta} + \frac{M}{1+\gamma}  \eta^{\gamma} \sigma_\gamma^{1+\gamma}, 
\end{eqnarray} 
where the constant $M$ is defined by \cref{assump:holderCondition}. In particular, if we let $\eta = c_\gamma / K^{1/(1+\gamma)}$ for some constant $c_\gamma>0$, then 
\begin{align}
\min_{0\leq k \leq K-1} \mathbb{E} \| \nabla f(\wb^k)\|^2 = \mathcal{O}\left(\frac{1}{K^{\frac{\gamma}{1+\gamma}}} \right).
\end{align}
Furthermore, if we choose
$ c_\gamma = \frac{1}{\sigma_\gamma}\sqrt[1+\gamma]{\frac{1+\gamma}{\gamma M }[f(\wb^0) - f_*]}$, then
\begin{align}
\min_{0\leq k \leq K-1} \mathbb{E} \| \nabla f(\wb^k)\|^2 \leq  \frac{a_\gamma}{K^{\frac{\gamma}{1+\gamma}}},
\end{align}
where $a_\gamma = \sigma_\gamma \left( \sqrt[\gamma+1]{\frac{(1+\gamma)}{\gamma} M}\right) [f(\wb^0) - f_*]^{\frac{\gamma}{\gamma+1}}$. 
\end{theorem}
We note that there is no clear `best choice' for the learning rate in deep learning practice, various decaying stepsize rules including the choices of $\eta=\mathcal{O}(1/\sqrt{K})$ and $\eta=\mathcal{O}(1/K)$ are proposed and are commonly used (see. e.g. \citep{reddi2016stochastic},\citep[Section 1]{wu2018wngrad}).
In fact, when gradients have finite variance, %
the choice of $\eta = \mathcal{O}(1/\sqrt{K})$ will lead to the fastest decay (with respect to $K$) of the right-hand side of (26) at a rate $\mathcal{O}(1/\sqrt{K})$. However, when the gradients are heavy-tailed, Theorem \ref{thm:local_conv} shows that this choice will lead to a slower rate of $\mathcal{O}(K^{-\frac{\gamma}{2}})$. Instead, if we choose the stepsize $\eta = \mathcal{O}(1/ K^{1/(1+\gamma)})$, the size of the gradients will shrink at a faster rate $\mathcal{O}\left({K^{-\frac{\gamma}{1+\gamma}}} \right)$. In particular, Theorem \ref{thm:local_conv} indicates that if the tail index $\alpha$ is smaller, the upper bound for $\gamma$ gets smaller and the convergence gets slower.
We note that some observations supporting our result have already been reported in the literature, e.g., in \citep{ge2018rethinking} it has been empirically shown that the decay rate $1/K$ performed better that $1/\sqrt{K}$ in a deep learning setup, which we believe is a consequence of the heavy-tailed nature of the problem.

\hspace{3pt}

\noindent \textbf{Convergence to global optima.} Besides convergence to local minima, we have also provided finite-time guarantees for discrete-time dynamics \eqref{eqn:sgd_alpha} in terms of suboptimality with respect to the global minimum in \citep[Section 1.1]{nguyen2019non} as a function of the stepsize and the scale parameter. In particular, it was shown that the heavy-tailed system has a worse dependency on both $K$ and $\eta$ as compared to the Gaussian case, which is in line with Theorem~\ref{thm:local_conv}.
Besides, it is known that if the scale parameter $\sigma$ gets smaller, the dynamics admits a stationary distribution that will concentrate more and more on the global minimizer although reaching out to stationary would require an exponential number of steps in the dimension in the worst case.

\section{Experimental Methodology}
\label{sec:exps}

Before presenting our numerical results, we describe our experimental methodology regarding how we estimate the heavy-tailedness of the stochastic gradients. First, we discuss how we can compute the tail-index $\alpha$ based on a recent estimator proposed in \cite{mohammadi2015estimating}. Second, we describe the procedure proposed in \citep{brcich2005stability} for testing whether stochastic gradients follow a symmetric $\alpha$-stable distribution.
\subsection{Tail index estimation}
Estimating the tail-index of an extreme-value distribution is a long-standing topic. Some of the well-known estimators for this task are \citep{hill1975simple,pickands1975statistical,dekkers1989moment,de1998comparison}. Despite their popularity, these methods are not specifically developed for $\alpha$-stable distributions and it has been shown that they might fail for estimating the tail-index for $\alpha$-stable distributions \citep{mittnik1996tail,paulauskas2011once}. 

In this section, we use a relatively recent estimator proposed in \citep{mohammadi2015estimating} for $\alpha$-stable distributions. It is given in the following theorem. 
\begin{theorem}[\cite{mohammadi2015estimating}]
Let $\{X_i\}_{i=1}^K$ be a collection of random variables with $X_i \sim \sas(\sigma)$ and $K = K_1 \times K_2$.
Define $Y_i \triangleq \sum_{j=1}^{K_1} X_{j+(i-1)K_1} \>$ for $i \in \llbracket 1, K_2 \rrbracket$. Then, the estimator
\begin{align}
\label{eqn:alpha_estim}
\widehat{\phantom{a}\frac1{\alpha}\phantom{a}} \hspace{-4pt} \triangleq \hspace{-2pt} \frac1{\log K_1} \Bigl(\frac1{K_2 } \sum_{i=1}^{K_2} \log |Y_i|  - \frac1{K} \sum_{i=1}^K \log |X_i| \Bigr). 
\end{align}
converges to $1/{\alpha}$ almost surely, as $K_2 \rightarrow \infty$.
\end{theorem}
As shown in Theorem 2.3 of \citep{mohammadi2015estimating}, this estimator admits a faster convergence rate and smaller asymptotic variance than all the aforementioned methods.

\begin{wrapfigure}{r}{0.4\columnwidth}
    \centering
    \vspace{4pt}
    \label{fig:exp_synth}   \includegraphics[width=0.32\columnwidth]{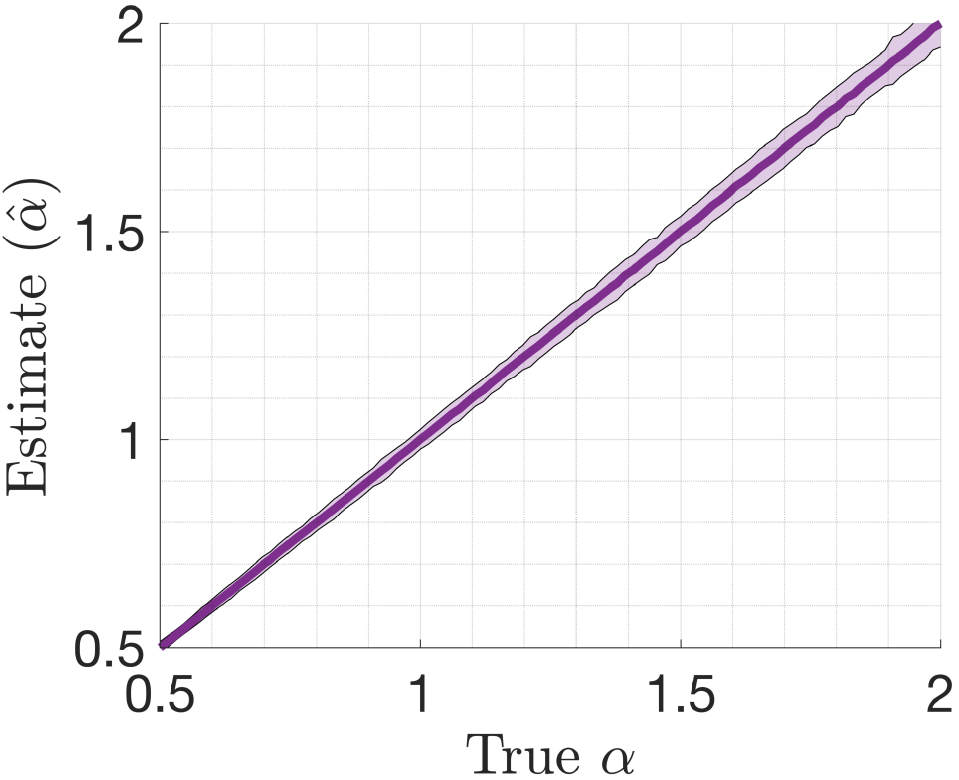}
    \caption{ Illustration of the tail-index estimator $\hat{\alpha}$. 
    \vspace{-2pt}
    }
\end{wrapfigure}
In order to verify the accuracy of this estimator, we conduct a preliminary experiment, where we first generate $K = K_1 \times K_2$ many $\sas(1)$ distributed random variables with $K_1= 100$, $K_2 =1000$ for $100$ different values of $\alpha$. 
Then, we estimate $\alpha$ by using $\hat{\alpha} \triangleq (\widehat{\phantom{a}\frac1{\alpha}\phantom{a}})^{-1}$. We repeat this experiment $100$ times for each $\alpha$. As shown in Figure~\ref{fig:exp_synth}, the estimator is very accurate for a large range of $\alpha$. Due to its favorable theoretical properties such as independence of the scale parameter $\sigma$, combined with its empirical stability, we choose this estimator in our experiments.

In order to estimate the tail-index $\alpha$ at iteration $k$, we first partition the set of data points $\mathcal{D} \triangleq \{1,\dots,n\}$ into many disjoint sets $\Omega_k^{i} \subset \mathcal{D}$ of size $b$, such that the union of these subsets give all the data points. Formally, for all $i,j =1,\dots, n/b$, $|\Omega_k^i| = b$, $\cup_{i} \Omega_k^i = \mathcal{D}$, and $\Omega_k^i \cap \Omega_k^j=\emptyset$ for $i \neq j$. This approach is similar to sampling without replacement. We then compute the full gradient $\nabla f(\wb_k)$ and the stochastic gradients $\nabla \tilde{f}_{\Omega_k^i}(\wb_k)$ for each minibatch $\Omega_k^i$. We finally compute the stochastic gradient noises $U^i_k(\wb_k) = \nabla \tilde{f}_{\Omega_k^i}(\wb_k) - \nabla f(\wb_k)$, vectorize each $U^i_k(\wb_k)$ and concatenate them to obtain a single vector, and compute the reciprocal of \eqref{eqn:alpha_estim}. In this case, we have $K=dn/b$ and we set $K_1$ to the divisor of $K$ that is the closest to $\sqrt{K}$.

\subsection{Stability test}\label{sec:stableTest}
Besides estimating the tail-index of a random process, it is also important to verify whether the process is symmetric $\alpha$-stable. In this section, we describe a procedure (\cite{brcich2005stability}) for obtaining a confidence level for the stability of a random process, based on the following property: 
\begin{theorem}[\cite{brcich2005stability}]
A necessary and sufficient condition for a random variable $X$ to have an $\sas$ distribution is
\begin{align}
    \label{eqn:stableTest_1}X_1 + X_2 &\sim C_1 X\\
    \label{eqn:stableTest_2}X_1 + X_2 + X_3 &\sim C_2 X
\end{align}
where $C_1,C_2>0$ and $X_1,X_2$ and $X_3$ are independent copies of $X$.
\end{theorem}
Here we adopt the stability test presented in \citep{brcich2005stability}. To obtain a statistical test from \eqref{eqn:stableTest_1}, we first separate the observations into three equal-size subsets $X$, $X_1$ and $X_2$, which are considered as independent copies of the observations. We then assign the first subset $X$ to the right side of \eqref{eqn:stableTest_1} and estimate the tail index $\alpha_X$ of this subset using the idea of the previous section. For the left side of \eqref{eqn:stableTest_1}, we sum $X_1$ and $X_2$ term by term, and estimate $\alpha_{12}$ of the resulting sum. Similarly, by separating the observations into four equal-size subsets $X'$, $X'_1$, $X'_2$ and $X'_3$, then repeating these above steps, we get $\alpha_{X'}$ from $X'$ and $\alpha_{123}$ from $X'_1 + X'_2 + X'_3$, for a statistical test of \eqref{eqn:stableTest_2}. In the end, the process is considered to be $\alpha$-stable if the tail indices estimated from the left and the right sides of \eqref{eqn:stableTest_1} (as well as of \eqref{eqn:stableTest_2}) are relatively close to each other, i.e. if the `condition number' $c_{st}\triangleq\max\{\vert\alpha_X-\alpha_{12}\vert,\vert\alpha_{X'}-\alpha_{123}\vert\}$ is smaller than some threshold.

\section{Results}\label{sec:experimentResult}

We investigate the tail behavior of the stochastic gradient noise in a variety of scenarios. We first consider a fully-connected network (FCN) on the MNIST and CIFAR10 datasets. For this model, we vary the depth (i.e.\ the total number of layers) in the set $\{2,3,\dots,10\}$, the width (i.e.\ the number of neurons per hidden layer) in the set $\{2,4,8,\dots,1024\}$, and the minibatch size ranging from $1$ to full batch. 

We then consider a convolutional neural network (CNN) architecture (AlexNet) on the CIFAR10 and CIFAR100 datasets. We scale the number of filters in each convolutional layer in range $\{2,4,\dots,512\}$. We use the existing random split of the MNIST dataset into train and test parts of sizes $60$K and $10$K, and CIFAR10 and CIFAR100 datasets into train and test parts of sizes $50$K and $10$K, respectively. The order of the total number of parameters $d$ range from several thousands to tens of millions.

For both FCN and CNN, we run each configuration with the negative-log-likelihood (i.e.\ cross entropy) and with the linear hinge loss, and we repeat each experiment with three different random seeds (see \citep{Geiger18} for details on the choice of the hinge loss). The training algorithm is SGD with no explicit modification such as momentum or weight decay. The training runs for a fixed number of iterations unless it hits 100\% training accuracy first. At every $100$th iteration, we log the full training and test accuracies, and the tail estimate of the gradients that are sampled using the corresponding mini-batch size. The codebase is implemented in python using pytorch \footnote{The codebase can be found at \url{https://github.com/umutsimsekli/sgd_tail_index}.}. %

Below, we present the most relevant and representative results. We have observed that, in all configurations, the three different initializations yielded no significant difference. Therefore, the effects of the randomness in initialization (under a given scheme) do not appear to affect the gradient noise. Similarly, the choice of the loss function do not yield different behaviours in terms of the tail index. Even though the heavy tailed nature remains the same, the choice of the loss function results in a different way of dependence to the hyperparameters of the system, which we discuss in Section~\ref{sec:gen} and leave the investigation to a further study.

\begin{figure}[t]
    \centering
    \subfigure[MNIST]{
    \label{fig:stableTest_fc_widths_layerwise_mnist}
    \includegraphics[width=0.32\columnwidth]{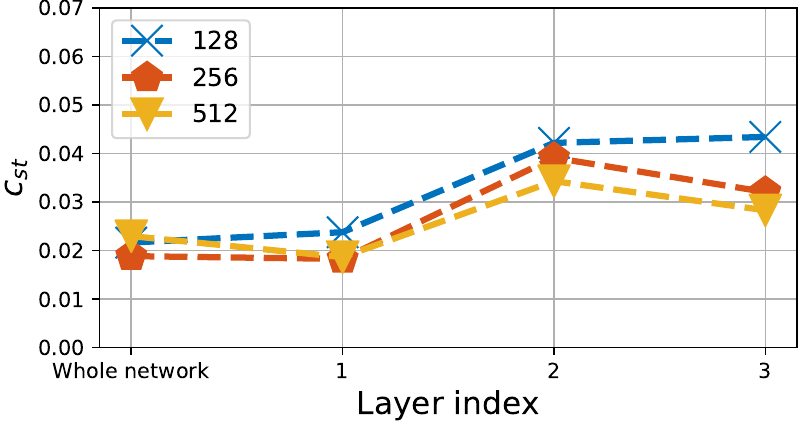}
    \includegraphics[width=0.32\columnwidth]{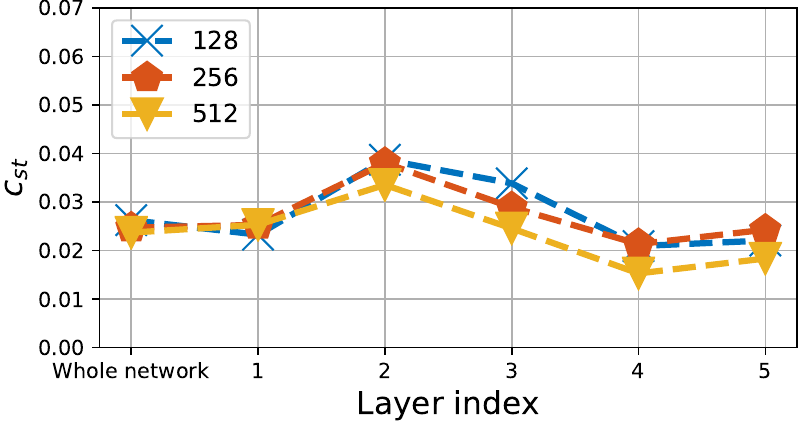}
    \includegraphics[width=0.32\columnwidth]{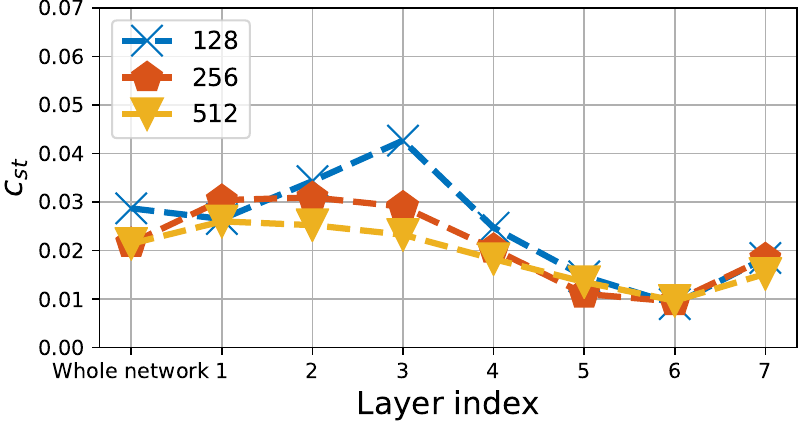}
    }
    \subfigure[CIFAR10]{
    \label{fig:stableTest_fc_widths_layerwise_cifar10}
    \includegraphics[width=0.32\columnwidth]{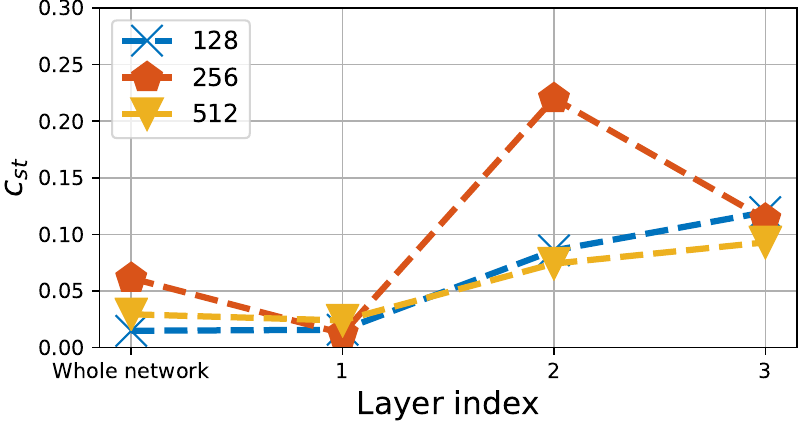}
    \includegraphics[width=0.32\columnwidth]{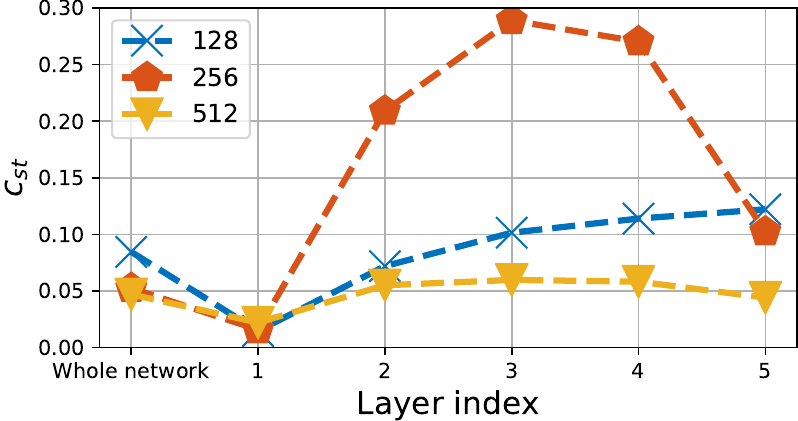}
    \includegraphics[width=0.32\columnwidth]{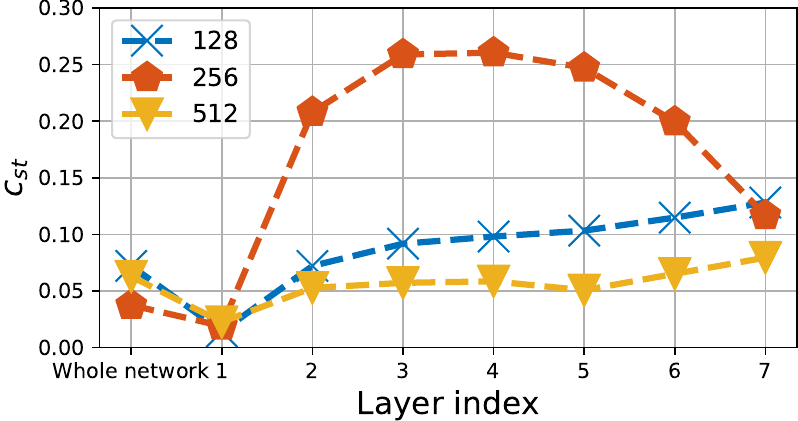}
    }
    \vspace{-10pt}
    \caption{Stability confidence results for each layer as well as for whole network (indicated by layer index at 0). From left to right: depth 3, depth 5, depth 7.   
}
    
\end{figure}

\subsection{Stability test results}

\label{sec:stability_test_res}

We first start by investigating the stability of the stochastic gradient noises under the datasets that we use for our estimation experiments. 
We will first focus on the later iterations of SGD, where the tail-index becomes stationary.
Using an FCN on the MNIST and CIFAR10 datasets, we estimate the condition number $c_{st}$ (as described in Section~\ref{sec:stableTest}) at every $50$th iteration of the training stage, then take its average over the last $10$K iterations to get the final result.
Here, we consider $c_{st} \approx 0.05$ to be an acceptable level for the test since it is a quite small number with respect to estimated $\alpha$ in our experiments.

The results using the MNIST dataset are illustrated by Figure~\ref{fig:stableTest_fc_widths_layerwise_mnist}, in which layer index at $0$ corresponds to the whole network while the indices $1,2,\ldots,7$ represent the hidden layers of the network. Our experiments show that the condition number $c_{st}$ for the whole network are always smaller than the threshold $0.05$, which means the gradient noise of the network satisfies our required stability criterion, even when we change the number of layers (depths) and the number of neurons per layer (widths). The same conclusion on the stability test is true when we investigate each of the hidden layers of the network.

Figure~\ref{fig:stableTest_fc_widths_layerwise_cifar10} shows the results of the stability test for CIFAR10. As can be seen from the figure, the condition $c_{st}$ of the network fails to be smaller than our required criterion in some cases. However, the gap from this number to the criterion is quite small that we can consider that it does not violate the $\alpha$-stable assumption on the gradient noise of the network. Unlike the MNIST dataset, we observe that for the networks with $256$ neurons per layer, even though the overall gradient noise strongly exhibits an $\alpha$-stable behavior, some of the hidden layers are very far from being $\alpha$-stable, suggesting that the characteristics of the first layer is dominating the overall structure. In contrast, the gradient noise with respect to the parameters of the hidden layers becomes more $\alpha$-stable with a very high number ($512$) of neurons per layer.

By these experiments, we observe that the structure of the dataset has a strong impact on the statistical properties of the gradient noise, especially for the layers with smaller number of parameters. When this number of parameters is large (which is usually the case in practice), the gradient noise corresponding to these parameters becomes more $\alpha$-stable. In short, this means increasing the size of the network (the number of the network parameters) tends to make the gradient noise behave similarly to an $\alpha$-stable noise.
\begin{figure}[t]
    \centering
    \subfigure[MNIST]{
    \includegraphics[width=0.24\columnwidth]{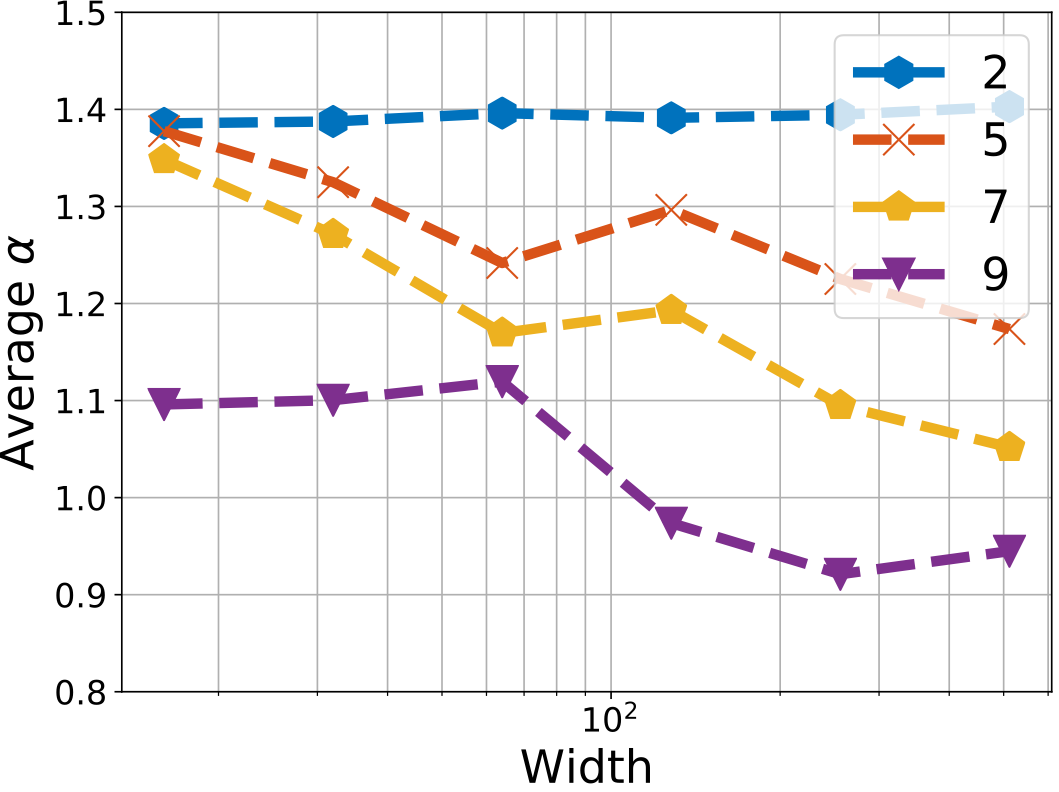}
    \includegraphics[width=0.24\columnwidth]{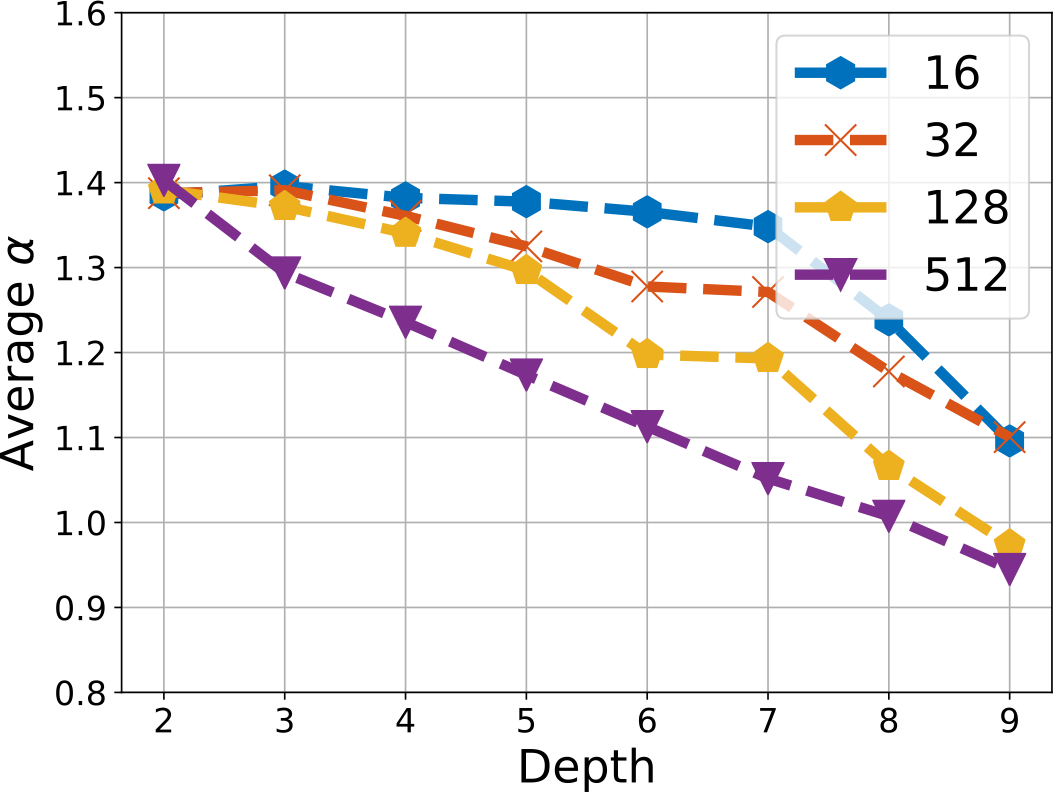}
    }
    \subfigure[CIFAR10]{\includegraphics[width=0.24\columnwidth]{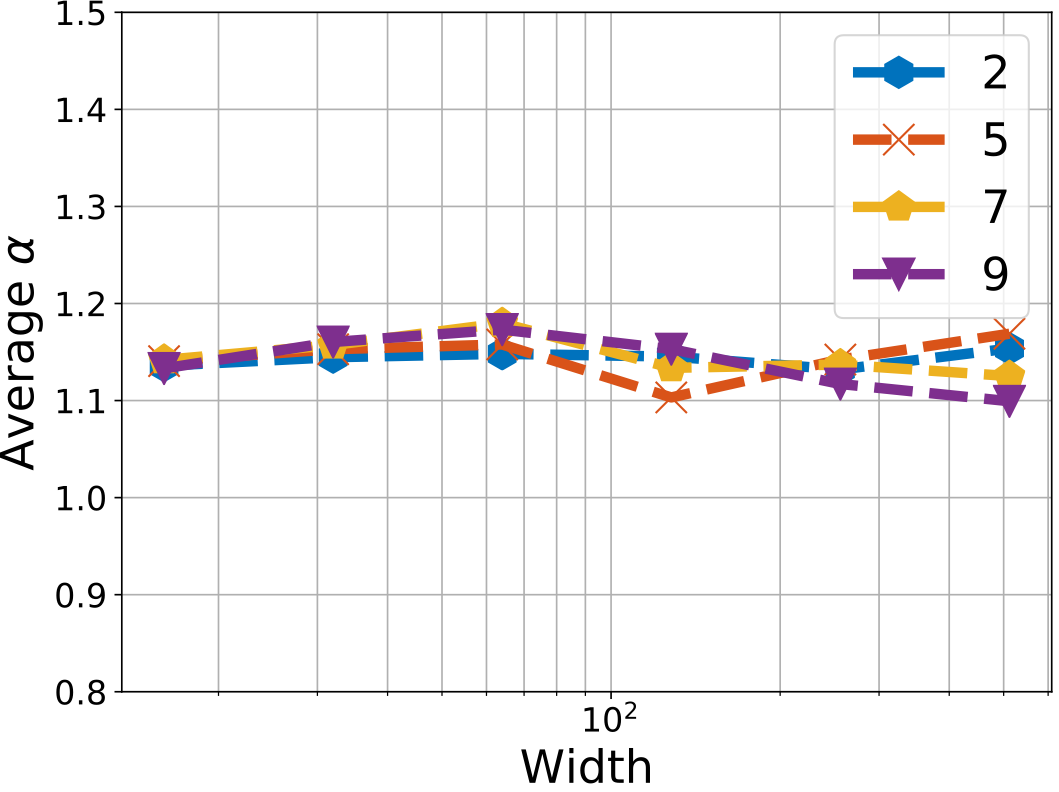}
    \includegraphics[width=0.24\columnwidth]{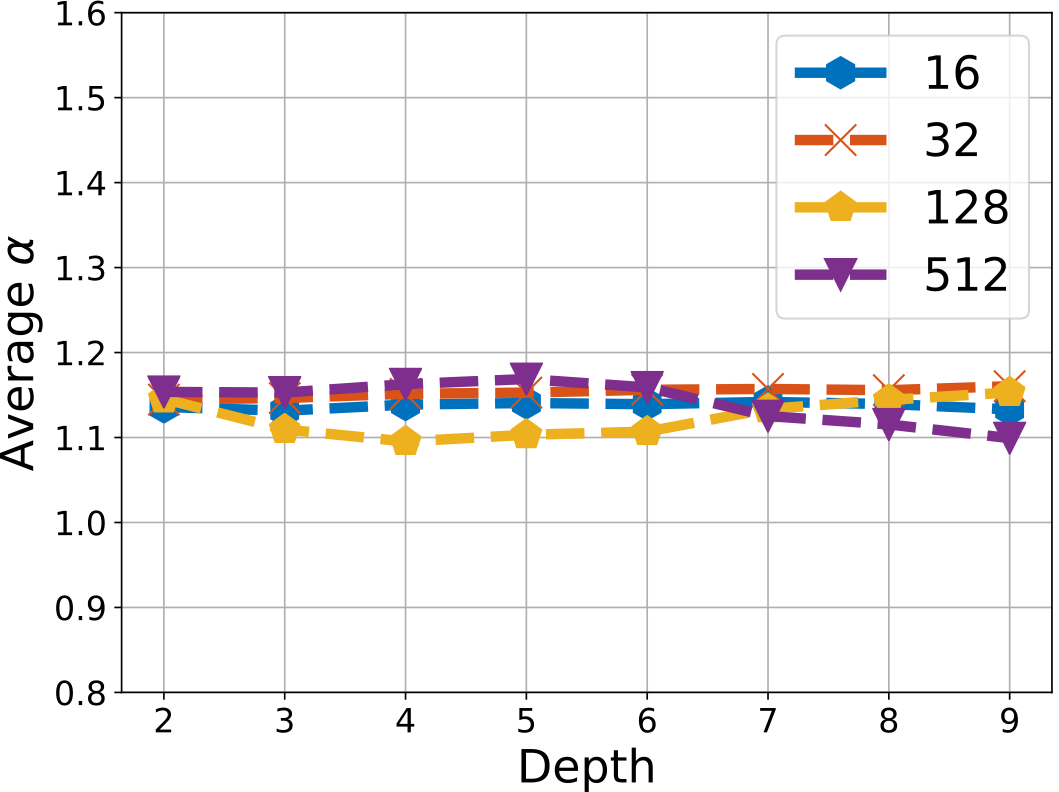}
    }
    \caption{Estimation of $\alpha$ for varying widths and depths in FCN. The curves in the left figures correspond to different depths, and the ones on the right figures correspond to widths.}
    \label{fig:exp_fc_widths}
\end{figure}

\begin{figure}[t]
    \centering
    \includegraphics[width=0.325\columnwidth]{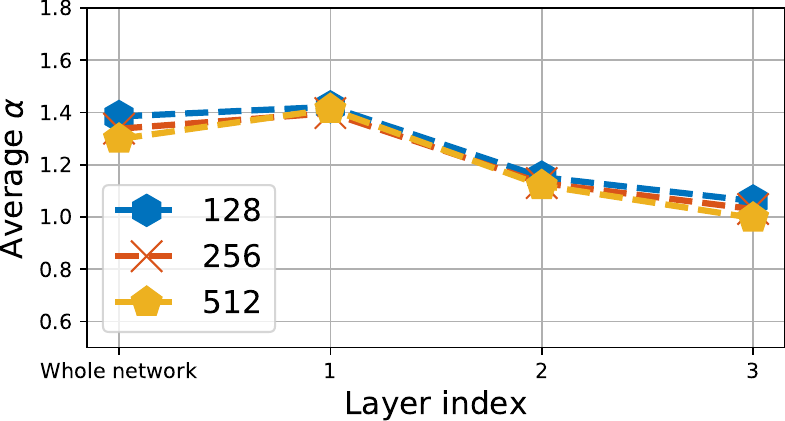}
    \includegraphics[width=0.325\columnwidth]{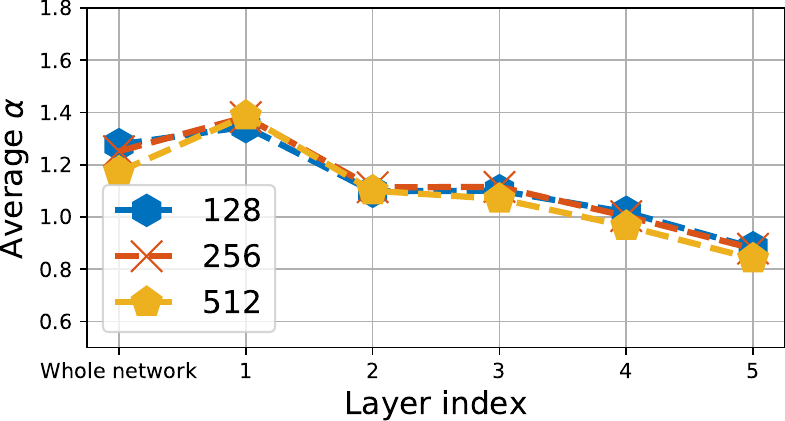}
    \includegraphics[width=0.325\columnwidth]{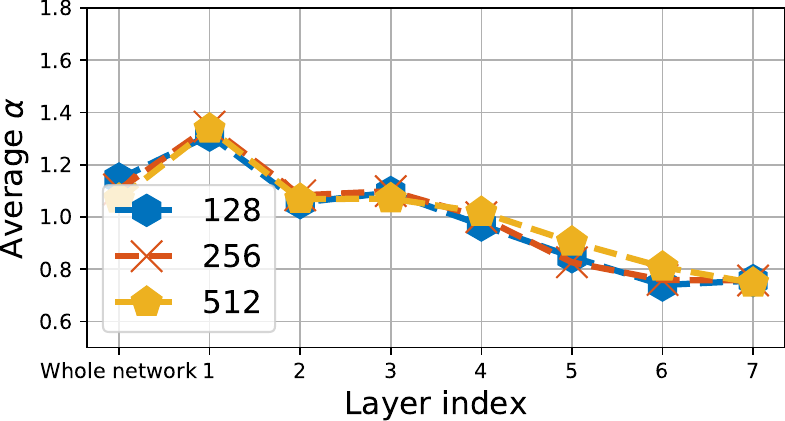}
    \caption{Estimation of $\alpha$ for varying widths and depths in FCN, dataset MNIST. From left to right: depth 3, depth 5, depth 7. Different lines correspond to different widths.}
    \label{fig:exp_fc_widths_layerwise_mnist}
\end{figure}

\subsection{Effect of varying network size}
\label{sec:exps_size}

We measure the tail-index for varying the widths and depths for the FCN, and varying widths (i.e.\ the number of filters) for the CNN. For very small sizes, the networks perform poorly; therefore, we only illustrate sufficiently large network sizes, which yield similar accuracies. For these experiments, we compute the average of the tail-index measurements for the last $10$K iterations (i.e.\ when $\hat{\alpha}$ becomes stationary) to focus on the late stage dynamics.

Figure~\ref{fig:exp_fc_widths} shows the results for the FCN. The first striking observation is that in all the cases, the estimated tail-index is far from $2$, meaning that the distribution of the gradient noise is highly non-Gaussian. For the MNIST dataset, we observe that $\alpha$ systematically decreases for increasing network size, where this behavior becomes more prominent with the depth. This result shows that, for MNIST, increasing the dimension of the network results in a gradient noise with heavier tails and therefore increases the probability of ending up in a wider basin. For the CIFAR10 dataset, we still observe that $\alpha$ is far from $2$; however, in this case, increasing the network size does not have a clear effect on $\alpha$. In all cases, we observe that $\alpha$ is in the range $1.1$--$1.2$.

In Figure~\ref{fig:exp_fc_widths_layerwise_mnist}, we plot estimated $\alpha$ for each layer of FCNs, using MNIST dataset where the minibatch is of size $100$. The resulting $\alpha$ is obtained by averaging $\alpha$ over the last $10$K iterations. The layer index `$0$' corresponds to the estimated $\alpha$ of the whole network.
In this experiment, we observe that $\alpha$ becomes smaller (heavier-tailed) for the deeper layers. In addition, the value of the tail-index for the whole network has a strong connection with the first layers: the $\alpha$ for the whole network is closer to that of the first layers than of the last layers.

\begin{figure}[t!]
    \centering
    \subfigure[CIFAR10 ]{
                    \includegraphics[width=0.24\columnwidth]{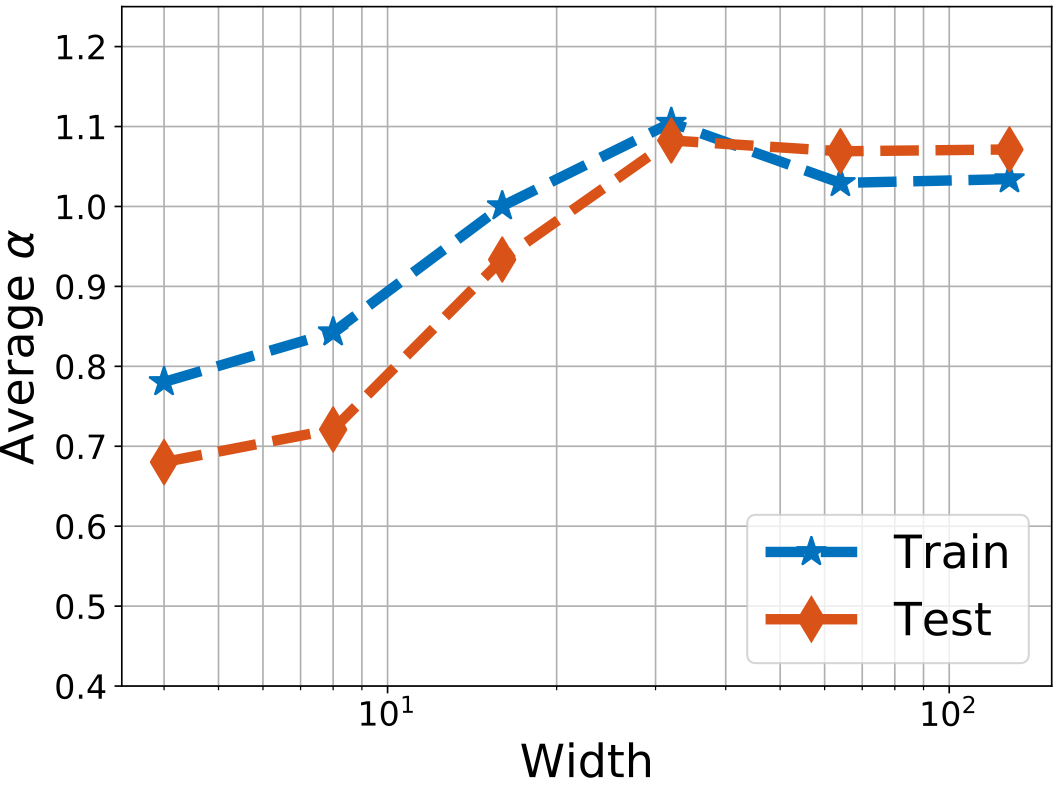}
                    \includegraphics[width=0.24\columnwidth]{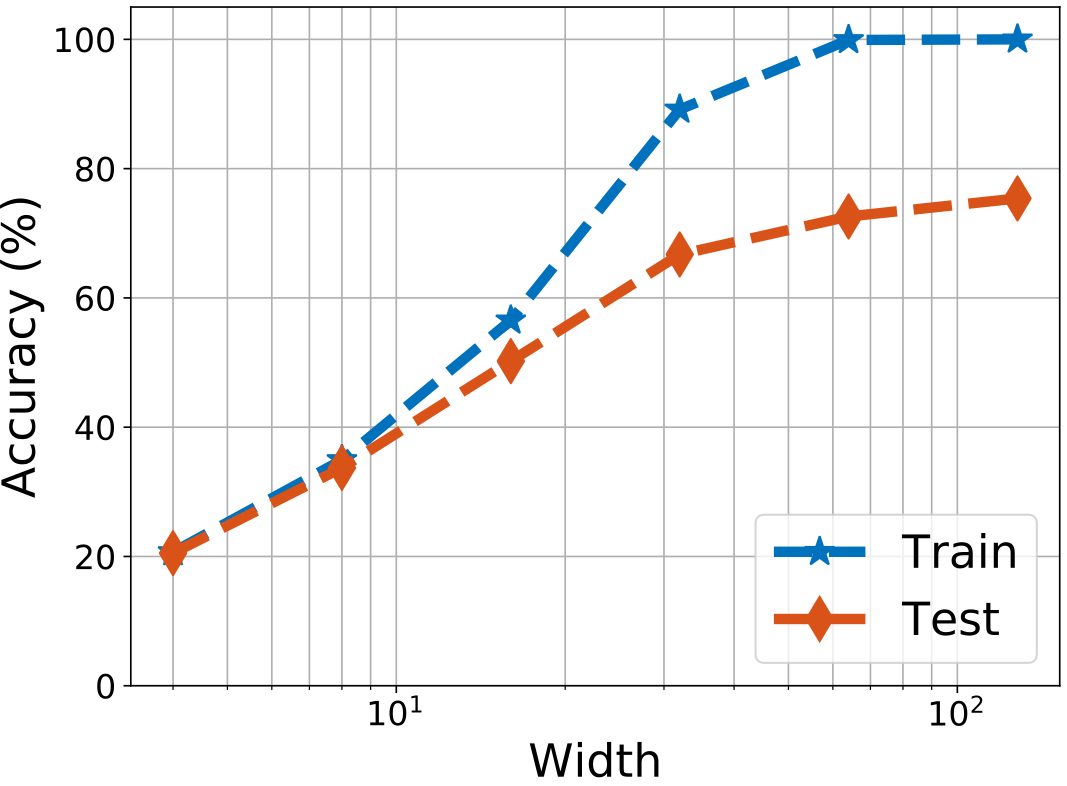}}
    \subfigure[CIFAR100 ]{
                    \includegraphics[width=0.24\columnwidth]{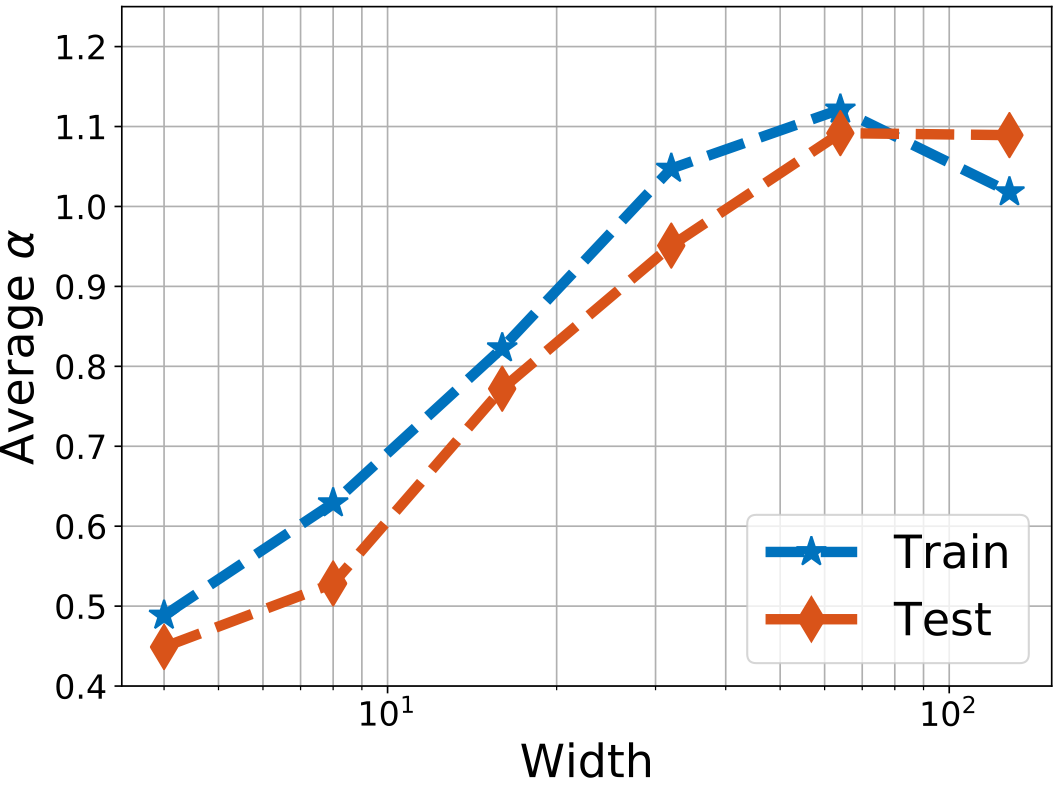}
                    \includegraphics[width=0.24\columnwidth]{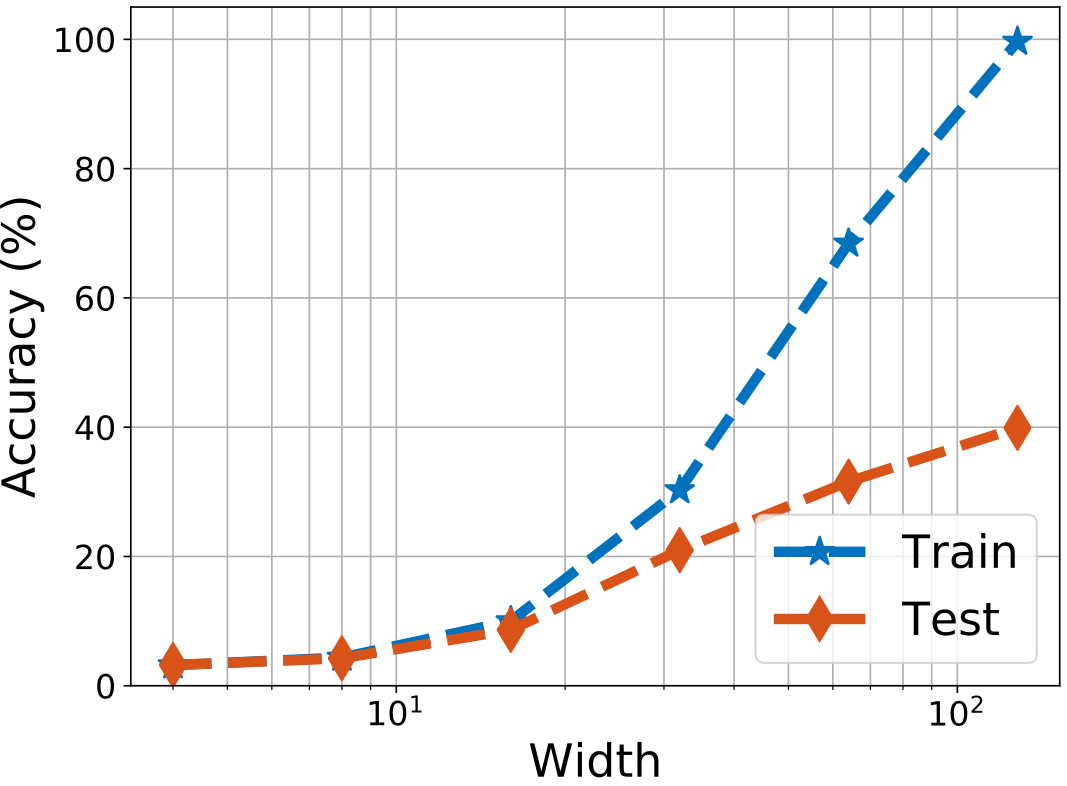}}
    \caption{The accuracy and $\hat{\alpha}$ of the CNN for varying widths.   }
    \label{fig:cnn_widths}
\end{figure}

Figure~\ref{fig:cnn_widths} shows the results for the CNN. In this figure, we also depict the train and test accuracy, as well as the tail-index that is estimated on the test set. These results show that, for both CIFAR10 and CIFAR100, the tail-index is extremely low for the under-parametrized regime (e.g.\ the case when the width is $2$, $4$, or $8$ for CIFAR10). As we increase the size of the network the value of $\alpha$ increases until the network performs reasonably well and stabilizes in the range $1.0$--$1.1$. We also observe that $\alpha$ behaves similarly for both train and test sets\footnote{We observed a similar behavior in under-parametrized FCN; however, did not plot those results to avoid clutter.}.

These results show that there is strong interplay between the network architecture, dataset, and the algorithm dynamics: (i) we see that the size of the network can strongly influence $\alpha$, (ii) for the exact same network architecture, the choice of the dataset has a significant impact on not only the landscape of the problem, but also the noise characteristics, hence on the algorithm dynamics. 

\subsection{Effect of the minibatch size}
\label{subsec:exps_mblr}

\begin{figure}[b]
    \centering
    \subfigure[Depth = 2]{\includegraphics[width=0.32\columnwidth]{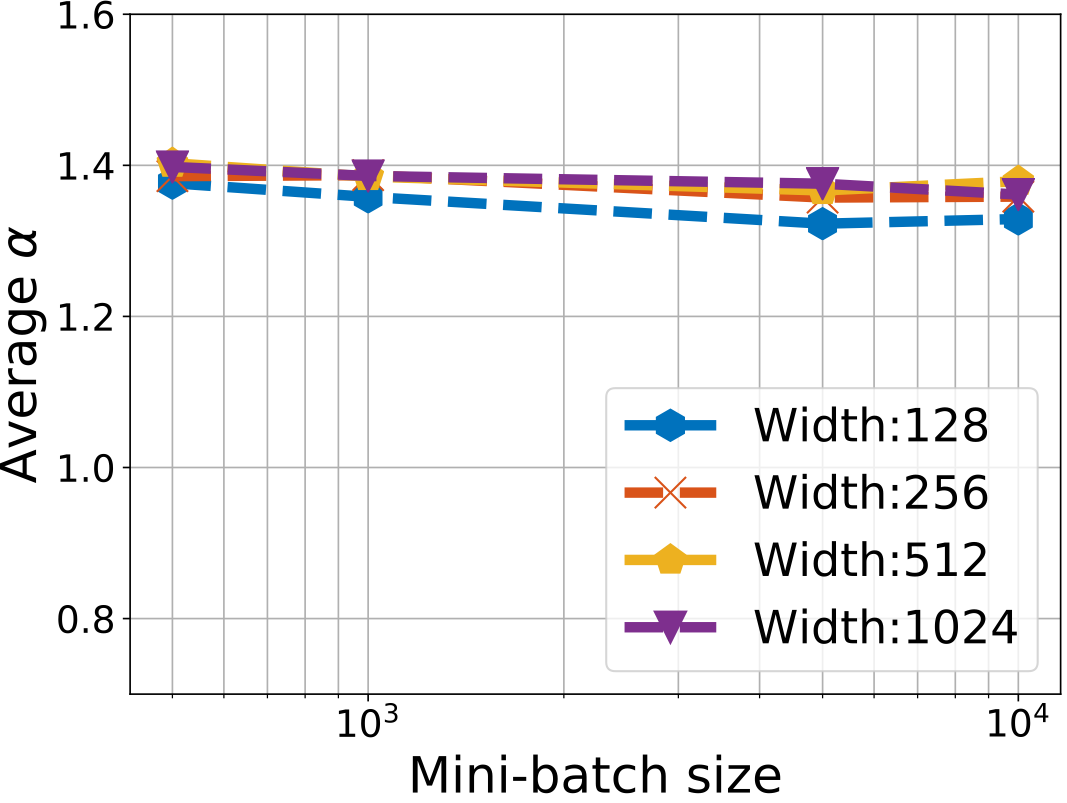}}
    \hspace{40pt}
    \subfigure[Depth = 4]{\includegraphics[width=0.32\columnwidth]{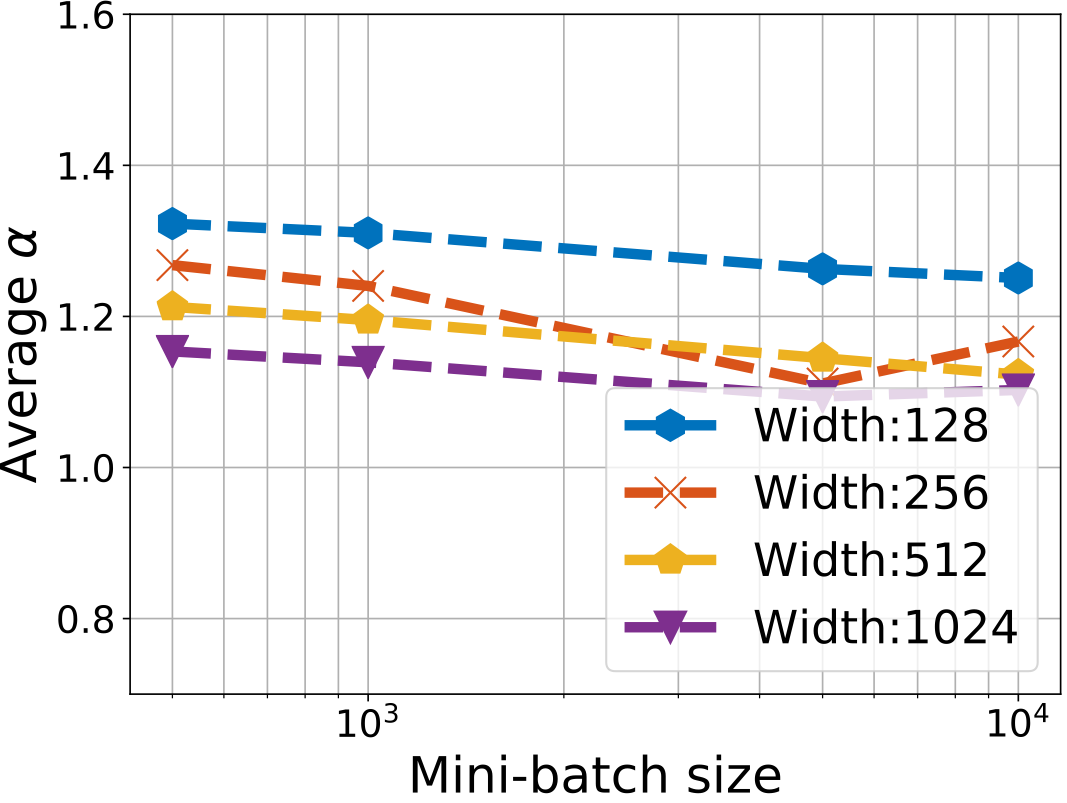}}
    \caption{Estimation of $\alpha$ for varying minibatch size.}
    \label{fig:exp_fc_mbscale}
\end{figure}

In our second set of experiments, we investigate the effect of the size of the minibatch on $\alpha$. We focus on the FCN and monitor the behavior of $\alpha$ for different network and minibatch sizes $b$. Figure~\ref{fig:exp_fc_mbscale} illustrates the results. 
This result contradicts with the Gaussian assumption of GN for large $b$, as the tail-index does not increase at all with the increasing batch size.
We observe that $\alpha$ stays almost the same when the depth is $2$ and it moves in a small interval when the depth is set to $4$. We note that we obtained the same train and test accuracies for different minibatch sizes.

\newcommand{\tempsize}{0.27}
\begin{figure}[t]
    \centering
    \subfigure[MNIST]{
    \includegraphics[width=\tempsize\columnwidth]{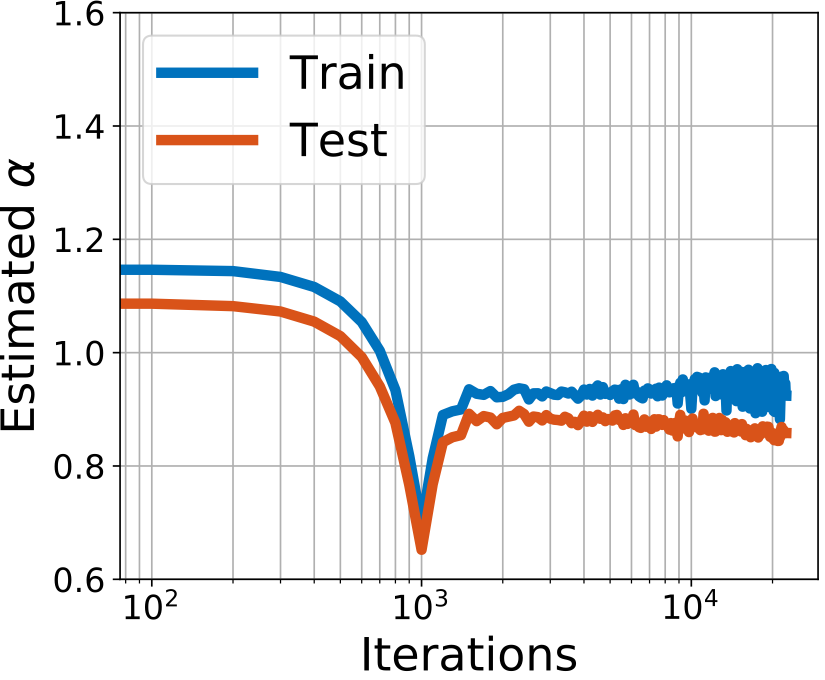}
    \includegraphics[width=\tempsize\columnwidth]{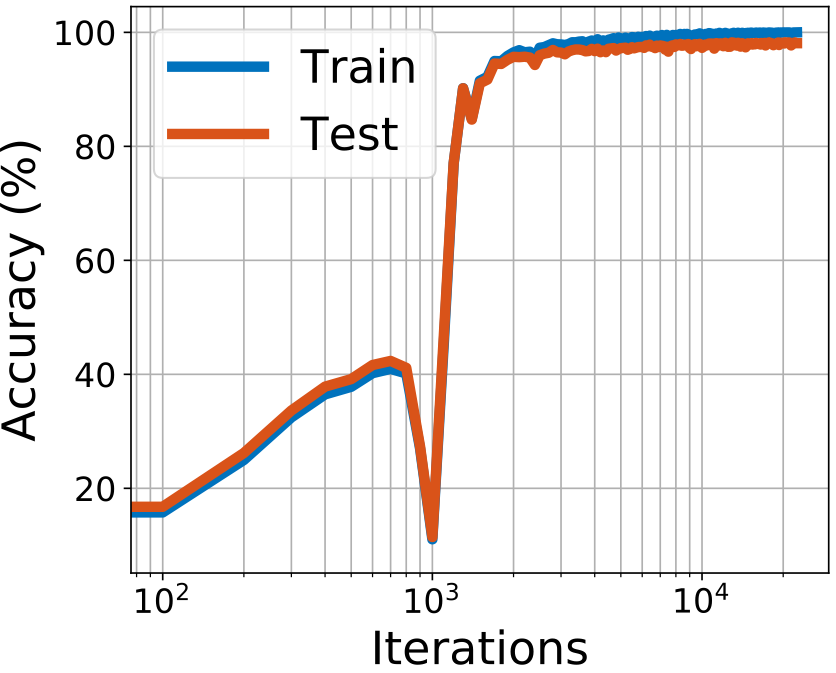}
    \includegraphics[width=\tempsize\columnwidth]{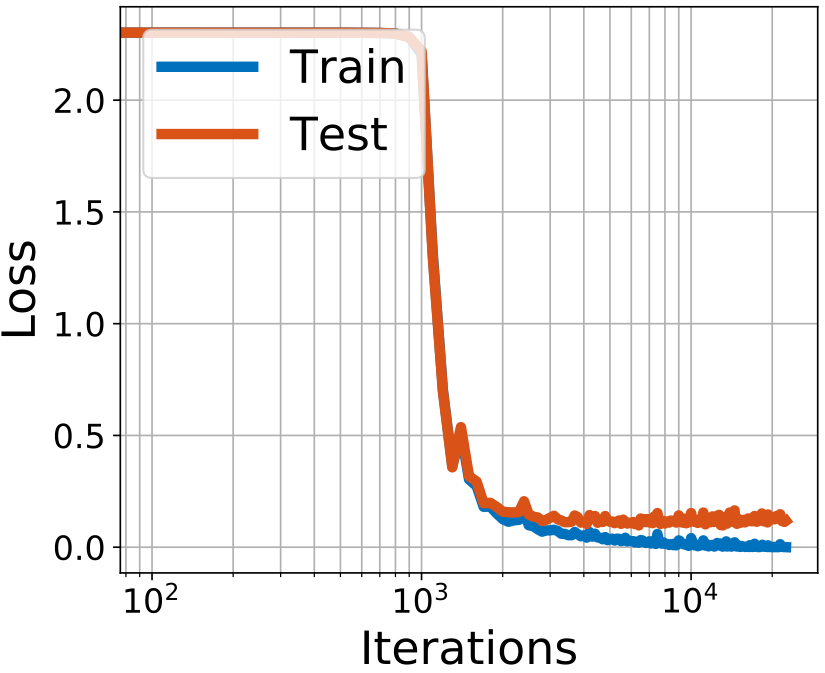}
    }
    \subfigure[CIFAR10]{
    \includegraphics[width=\tempsize\columnwidth]{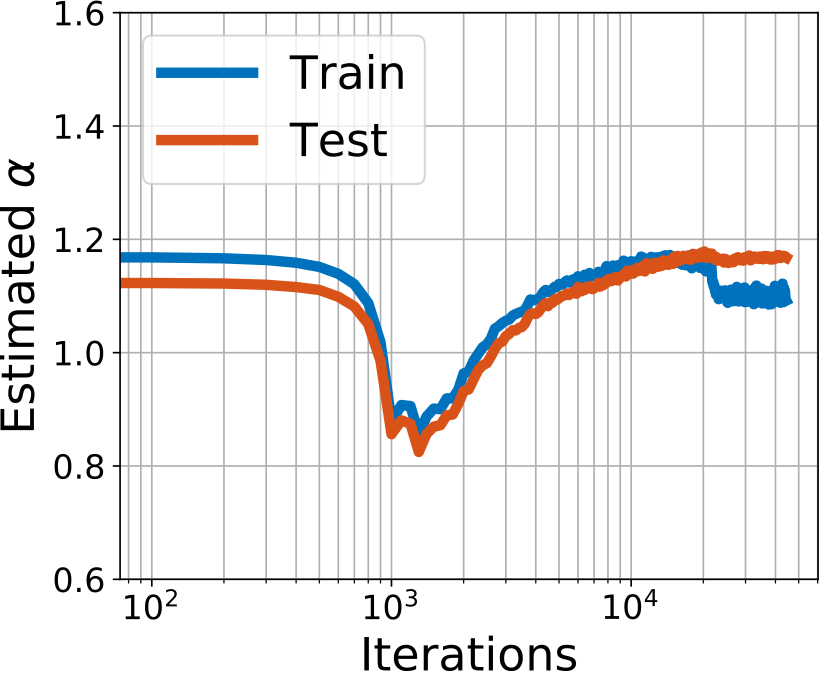}
    \includegraphics[width=\tempsize\columnwidth]{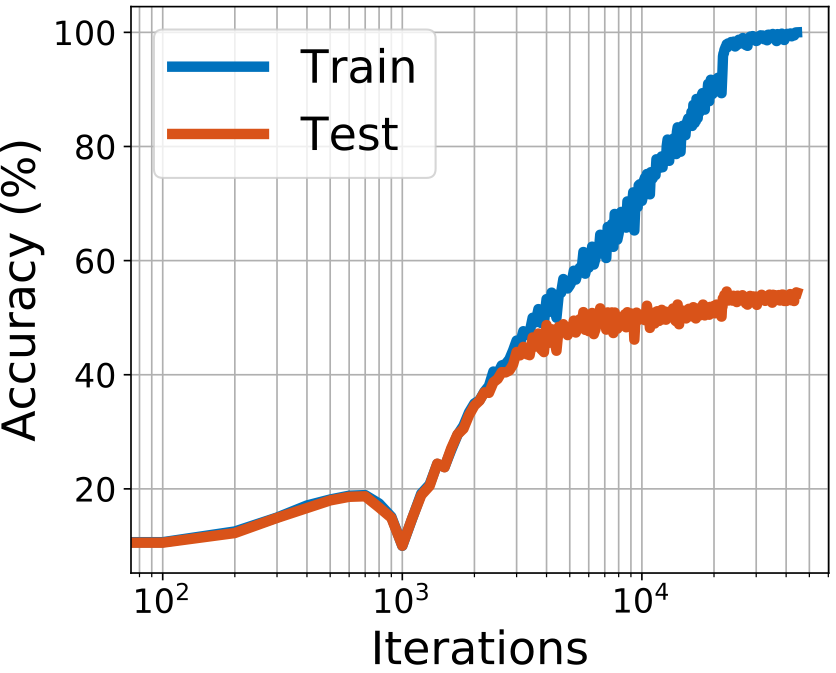}
    \includegraphics[width=\tempsize\columnwidth]{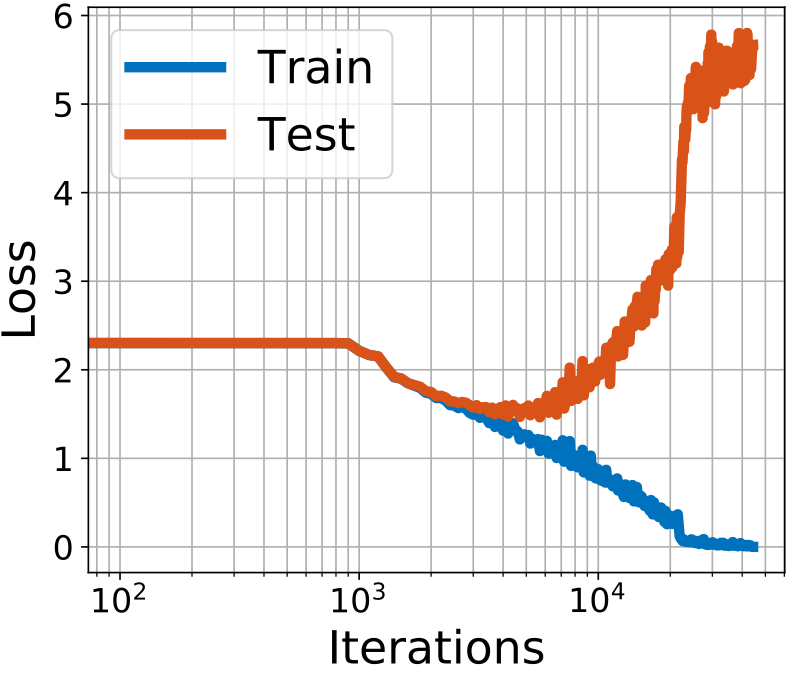}
    }
    \caption{The iteration-wise behavior of of $\alpha$ for the FCN.}
    \label{fig:exp_iter_fc}
\end{figure}

\begin{figure}[b]
    \centering
    \includegraphics[width=0.53\columnwidth]{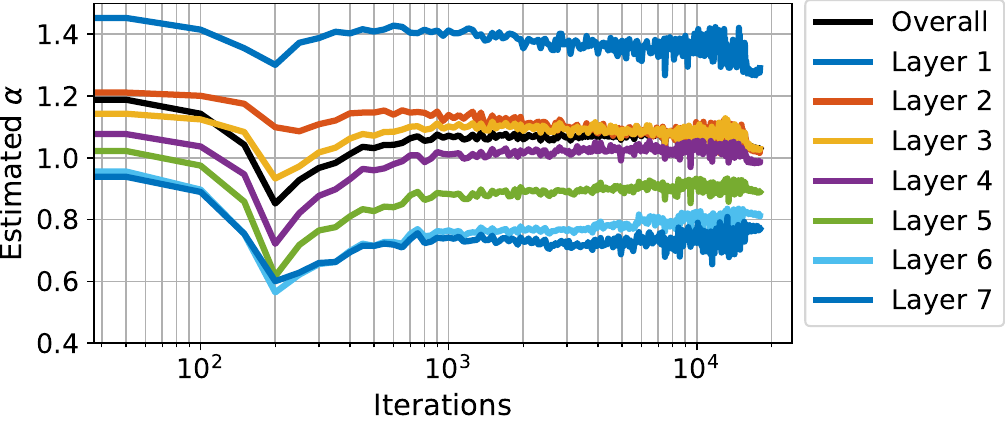}
    %
    %
    
    \caption{Estimation of $\alpha$ with an FCN on MNIST.}
    \label{fig:exp_fc_widths_layerwise_mnist_jump}
\end{figure}

\subsection{Tail behavior throughout the iterations}
\label{subsec:exp_iter}

So far, we have focused on the late stages of SGD, where $\alpha$ is in a rather stationary regime. In this set of experiments, we shift our focus on the first iterations and report an intriguing behavior that we observed in almost all our experiments. As a representative, in Figure~\ref{fig:exp_iter_fc}, we show the temporal evolution of SGD for the FCN with $9$ layers and $512$ neurons/layer. 

The results clearly show that there are two distinct phases of SGD (in this configuration before and after iteration $1000$). In the first phase, the loss decreases very slowly, the accuracy slightly increases, and more interestingly $\alpha$ rapidly decreases. When $\alpha$ reaches its lowest level, the process possesses a jump, which causes a sudden decrease in the accuracy. After this point the process recovers again and we see a stationary behavior in $\alpha$ and an increasing behavior in the accuracy.

We also investigate this behavior for each layer of an FCN with depth $7$ and width $512$ in Figure~\ref{fig:exp_fc_widths_layerwise_mnist_jump}. The estimated tail-index for each layer has a clear phase change at earlier iterations, where we observe that this jump is more prominent in the deeper layers where the tail-index is smaller. 
On the other hand, unlike the whole network, the tail-index of each layer undergoes a fluctuation period before becoming stationary at the last $2000$ iterations. However, this observation might be due to the measurement error since the size of the sample that is used in the estimator \eqref{eqn:alpha_estim} gets smaller when we make layer-wise measurements.

The fact that the process has a jump when $\alpha$ is at its smallest value provides a strong support to our assumptions and the metastability theory that we discussed in the previous section. Furthermore, these results also strengthen the view that SGD crosses barriers at the very initial phase and continues searching until it reaches a ``wide and flat enough" region of a local optimum.  On the other hand, our current analysis is not able to determine whether the process jumps in a different basin or a `better' part of the same basin and we leave it as a future work.

\subsection{A note on generalization}
\label{sec:gen}

In this section, we investigate the connection between the tail-index and the generalization performance. In particular, we consider the relation between $\alpha$ and the ratio of the step-size to the batch-size $\eta/b$ which is proportional to the \emph{noise scale} of SGD when there is no momentum \citep{park2019effect}. It has been empirically demonstrated that this ratio correlates with performance of the model \citep{jastrzkebski2017three}, hence the higher the noise scale, the better the generalization performance until a certain level. Clearly, when the noise is too high, training may diverge, however, proper level of noise leads to better solutions. 

In this section, we will investigate how the tail index of the gradient noise is affected for different noise scales. We reproduce and follow the initialization convention and the hyper-parameter scale that is studied in \citep[Appendix G]{park2019effect}: A fully connected model with 3 hidden layers, each hidden layer has 512 nodes. Weights are initialized $\sim\mathcal{N}(0, 1)$, bias terms are set to zero at the initial point. Each layer, is then passed through ReLU non-linearity, and multiplied by the inverse of the width of the previous layer. As usual, the network is trained with SGD without momentum; the dataset is the standard MNIST. Minibatch size ranges in the set $[24,48,96,192]$ and step-size ranged from the set $[0.9375, 1.875, 3.75, 7.5, 15]$\footnote{Note that this particular scaling is introduced in \cite{jacot2018neural} and it admits slightly larger values of learning rates compared to standard initialization schemes}.

\begin{figure}[t]
    \centering
    \subfigure[Test error vs $\eta/b$]{
    \label{fig:lrBs_testError}
    \includegraphics[width=0.235\columnwidth]{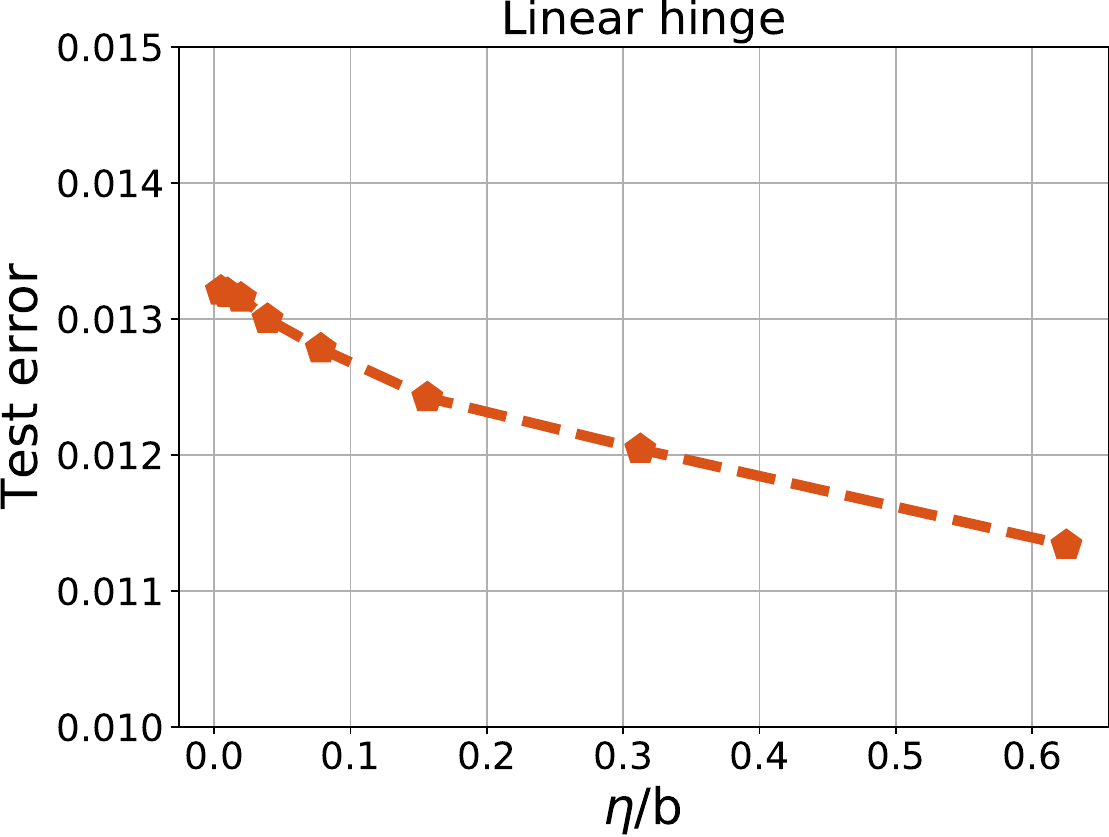}
    \includegraphics[width=0.235\columnwidth]{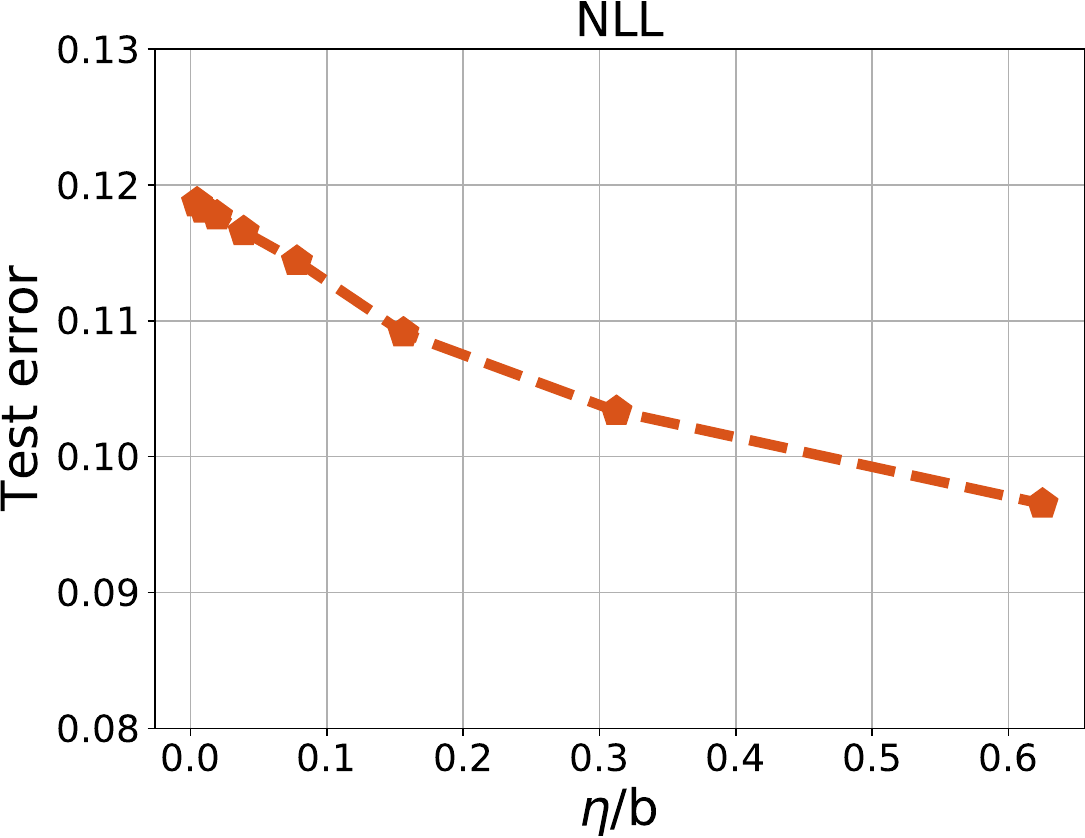}
    }
    \subfigure[Tail-index vs $\eta/b$]{
    \label{fig:lrBs_alpha}
    \includegraphics[width=0.235\columnwidth]{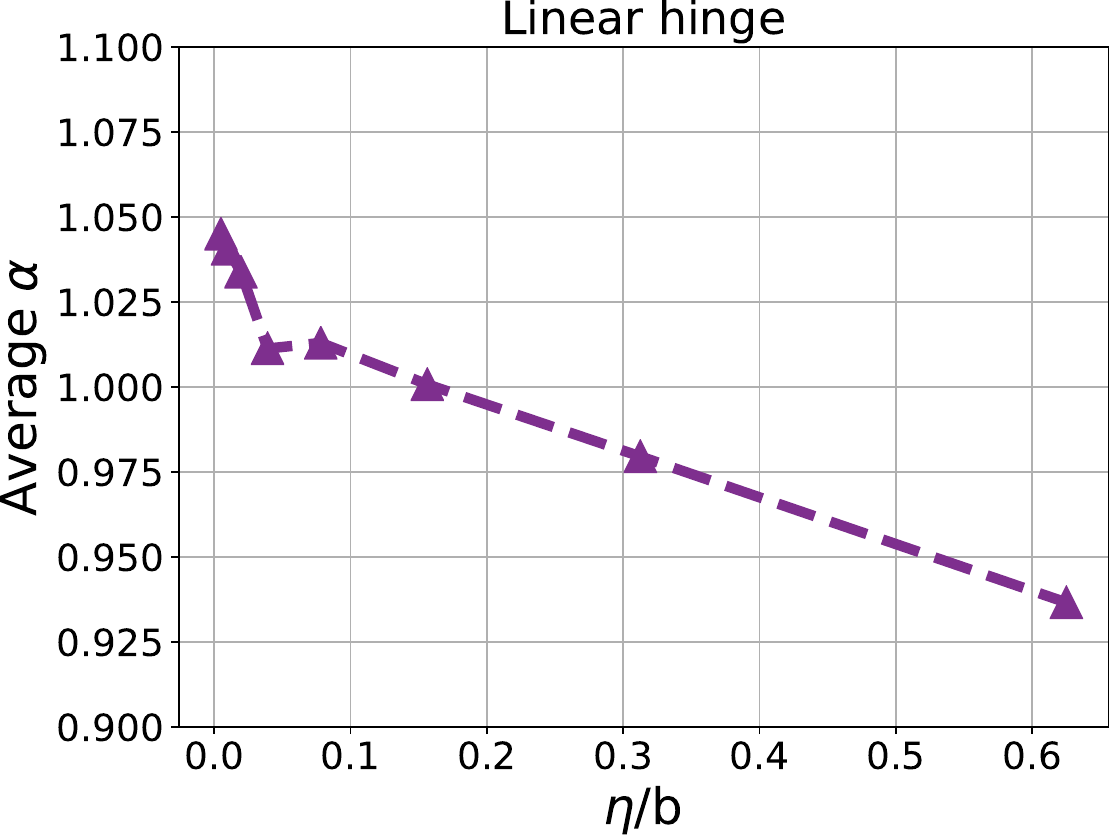}
    \includegraphics[width=0.235\columnwidth]{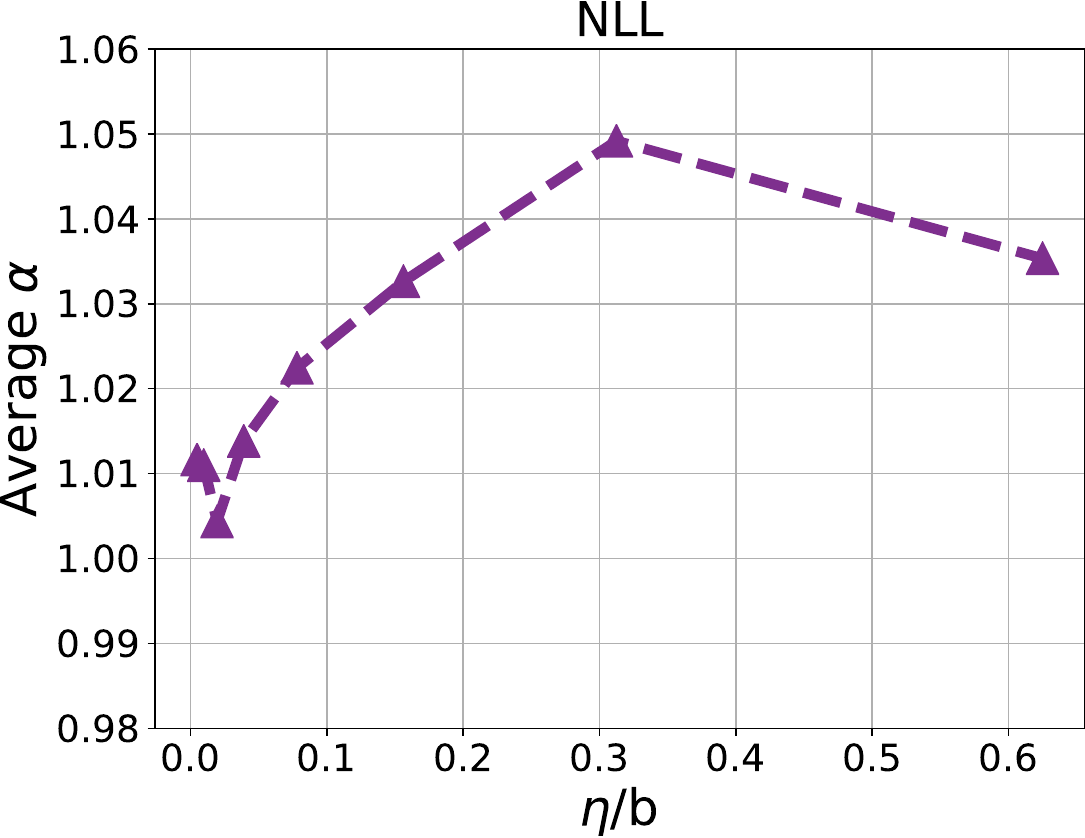}
    }
    \caption{Test error and tail-index in accordance with the change of $\eta/b$ ratio.}
\end{figure}

Figure~\ref{fig:lrBs_testError} and Figure~\ref{fig:lrBs_alpha} visualize the results. The estimated $\alpha$ and the test error are averaged among the candidates with the same $\eta/b$, at the last iteration of the training process. We ignore the particular values of test error but rather focus on the way certain choices affect the trends in the system and the behaviour of SGD dynamics.

\begin{figure}[t]
    \centering
    \subfigure[Linear hinge]{
    \includegraphics[width=0.45\columnwidth]{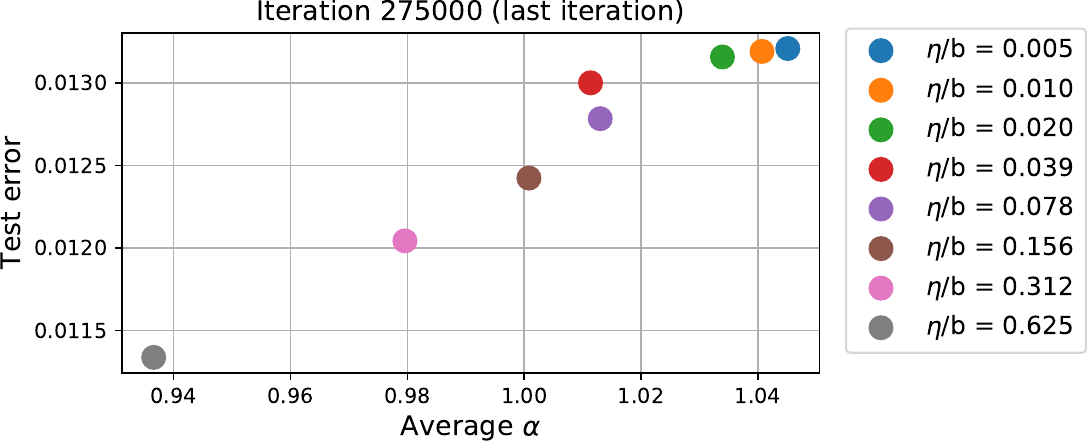}
    }
    \subfigure[NLL]{
    \includegraphics[width=0.45\columnwidth]{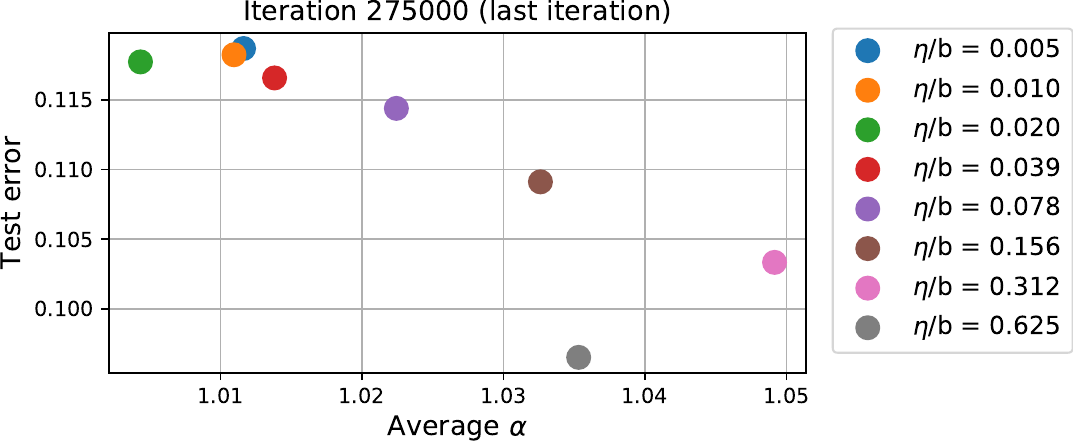}
    }
    \vspace{-10pt}
    \caption{Test error, estimated $\alpha$ in accordance with the change of $\eta/b$ ratio.}
    \label{fig:alpha_testError}
\end{figure}

In both choices of loss functions, hinge and NLL, the behaviour of the test error with respect to the noise scale is consistent with previous observations. Similarly, in both cases, the estimated $\alpha$ remains within a narrow band of 1, indicating the heavy tail behaviour. However, the trends in estimated $\alpha$ are different depending on the choice of the loss. Therefore, we cannot attribute the improvement in performance to lower $\alpha$ when increasing the noise scale. To better emphasize this point, we plot the correlation of estimated $\alpha$ and test error in Figure~\ref{fig:alpha_testError} where the positive and negative correlations are clearly visible depending on the choice of the loss function. This contrasting behaviour is another hint that there exists a connection between $\alpha$ and the test performance (since they are correlated in both cases) and suggests us to examine this connection in order to understand when exactly the dynamics falls into basins with better performance.

\section{Conclusion and Open Problems}
\label{sec:conc}

We investigated the tail behavior of the gradient noise in deep neural networks and empirically showed that the gradient noise is highly non-Gaussian. This outcome enabled us to analyze SGD as an SDE driven by a L\'{e}vy motion and establish a bridge between SGD and existing theoretical results, which provides more insights on the behavior of SGD, especially in terms of choosing wide minima. We also proved a new result on the local convergence of SGD, which provided justification to the step-size decay rates that are used in practice. 

Our study also brings up the following questions: 
\begin{itemize}[noitemsep,topsep=1pt,leftmargin=*,align=left]
    \item We observe that the tail-index might depend on the current state $\wb^k$, which suggests analyzing SGD as a `stable-like process' where the tail-index can depend on time \citep{bass1988uniqueness}. However, the metastability behavior of these processes are not clear at the moment and its theory is still in an early phase \citep{kuhwald2016bistable}.
    \item At the initial point, in the over-parametrized regime with large batch sizes, the noise can in fact be of Gaussian nature (cf.\ Figure~\ref{fig:noise_norms}). However, this property is destroyed quickly (see \cite{Neal1996,NIPS2005_2869,Lee2017} for a discussion on the infinite width networks, and \cite{panigrahi2019non} for a discussion on the early phases and large batches). We note that such Gaussianity heavily depends on the structure of the data, initialization scheme, and the size of the network in a sensitive way and may hold in only certain regimes or in specific cases. We think that identifying the crossover between the Gaussian and non-Gaussian regimes depending on the architecture and data is an important open problem.
    \item We have empirically observed a heavy-tailed behavior in the stochastic gradient noise; however, it is still not (rigorously) clear what the underlying mechanism that drives this heavy-tailed behavior is. Therefore, investigating this underlying mechanism would be an interesting future direction.
    \item Even though the general heavy-tailed behaviour remains unchanged with the choice of the loss function, we still observe different behaviours in terms of relation to generalization (see Figure~\ref{fig:alpha_testError}).
    We note that our results are related to the findings of \citep{martin2019traditional}, which modeled the weight matrices as heavy-tailed random matrices and investigated the density of the singular values of those matrices. Their empirical results on various different types of neural networks show that when the batch size gets smaller, the training process is able to catch finer-scale correlations from the data, leading to more strongly-correlated models between the layers of the network and that the entries of the weight matrices and the density of its singular values have heavier tails.  Our results in \ref{sec:gen} are partially consistent with the findings of \citep{martin2019traditional}. Their results combined with ours would shed more light into the heavy-tailedness of the SGD iterates and generalization properties of SGD algorithms. In the future, we would like to investigate the underlying deeper connections between the heavy-tailed behavior and generalization
    further from both a mathematical and experimental perspective.
    \item An extension of the current metastability theory that includes minima with zero modes is also missing. Such an extension would require redefining the set $A$ in the first exit time problem \eqref{eqn:set_A} and appears to be a challenging yet important direction of future research. 
\end{itemize}

\section*{Acknowledgments}

The contribution of authors Umut \c{S}im\c{s}ekli, Thanh Huy Nguyen, and Ga\"{e}l Richard to this work is partly supported by the French National Research Agency (ANR) as a part of the FBIMATRIX (ANR-16-CE23-0014) project, and by the industrial chair Data science \& Artificial Intelligence from T\'{e}l\'{e}com Paris. Mert G\"{u}rb\"{u}zbalaban acknowledges support from the grants NSF DMS-1723085 and NSF CCF-1814888.

\appendix

\section{Proof of Theorem~\ref{thm:local_conv}}

\begin{proof} We follow the proof technique of \citep[Theorem 1]{reddi2016stochastic} for the special case $\gamma=1$ and extend it to the more general case when $\gamma \leq 1$. Let $\mathcal{F}_k$ be the natural filtration associated to the algorithm up and including step $k$ and the random variable $\wb^k$. Let $\mathbb{E}_k$ denote the conditional expectation with respect to $\mathcal{F}_k$. 

By \cref{assump:holderCondition} which indicates that the gradient of the objective $f$ is H\"older with constant $\gamma$, and
by \citep[Lemma 1]{lei2019stochastic} and \citep[Lemma S2]{nguyen2019non}, we have 
 \begin{equation} f({\wb_2}) \leq f(\wb_1) + \langle \nabla f(\wb_1), \wb_2 - \wb_1\rangle + \frac{M}{1+\gamma}\|\wb_1 - \wb_2\|^{1+\gamma}, 
 \label{ineq}
 \end{equation}
for every $\wb_1, \wb_2 \in \mathbb{R}^d$, where $\langle \cdot, \cdot \rangle$ denotes the Euclidean inner product.

We first estimate $\mathbb{E}_k f(\wb^{k+1})$ by setting  $w_2 = \wb^{k+1}$ and $w_1 = \wb^k$ in \eqref{ineq}:
\begin{eqnarray}
\mathbb{E}_k f(\wb^{k+1})  &\leq& 
\mathbb{E}_k \left( f(\wb_k) + \langle \nabla f(\wb^k, \wb^{k+1}-\wb^k \rangle  + \frac{M}{1+\gamma} \eta^{1+\gamma}\|  \nabla \tilde{f}(\wb^k) \|^{1+\gamma}\right) \\
 &=& 
 f(\wb^k) - \eta \langle \nabla f(\wb^k), \mathbb{E}_k { \nabla \tilde{f}(\wb^k)}\rangle + \frac{M}{1+\gamma} \eta^{1+\gamma} \mathbb{E}_k {\|  \nabla \tilde{f}(\wb^k) \|^{1+\gamma}}\\
 &\leq &f(\wb^k) - \eta \| \nabla f(\wb^k)\|^2 + \frac{M}{1+\gamma} \eta^{1+\gamma}\sigma_\gamma^{1+\gamma},
 \end{eqnarray}
 where in the last step we used \cref{assump:holderCondition}. Taking expectations with respect to the random variable $\wb^k$, by the tower property of the expectations, we obtain
 $$\mathbb{E} f(\wb^{k+1}) \leq  \mathbb{E} f(\wb^k) - \eta \mathbb{E} \| \nabla f(\wb^k)\|^2 + \frac{M}{1+\gamma} \eta^{1+\gamma}\sigma_\gamma^{1+\gamma}.$$

Reorganizing the terms, 
$$ \mathbb{E}\| \nabla f(\wb_k)\|^2 \leq \frac{\mathbb{E} f(\wb^k) - \mathbb{E} f(\wb^{k+1})}{\eta} + \frac{M}{1+\gamma}  \eta^{\gamma} \sigma_\gamma^{1+\gamma}.
$$
Summing this inequality over $k$ from $0$ to $K-1$, we obtain
\begin{eqnarray}
\min_{0\leq k \leq K-1} \mathbb{E} \| \nabla f(\wb^k)\|^2 &\leq& \frac{1}{K} \sum_{k=0}^{K-1} \mathbb{E} \| \nabla f(\wb^k)\|^2 \\
& \leq & \frac{f(\wb^0) - f(\wb^k)}{K\eta} + \frac{M}{1+\gamma}  \eta^{\gamma} \sigma_\gamma^{1+\gamma} \\
& \leq & \frac{f(\wb^0) - f_*}{K\eta} + \frac{M}{1+\gamma}  \eta^{\gamma} \sigma_\gamma^{1+\gamma}. 
\end{eqnarray}
If we plug in $\eta = \frac{c_\gamma}{K^{1/(1+\gamma)}}$ in the last step, we obtain
\begin{eqnarray}
\min_{0\leq k \leq K-1} \mathbb{E} \| \nabla f(\wb^k)\|^2 & = & \frac{1}{K^{\gamma/(1+\gamma)}} \left( \frac{f(\wb^0) - f_*}{ c_\gamma } + \frac{M}{1+\gamma}  c_\gamma^{\gamma} \sigma_\gamma^{1+\gamma}\right). \label{ineq-to-min}
\end{eqnarray} 
It follows after a straightforward computation that the choice of $c_\gamma = \frac{1}{\sigma_\gamma}\sqrt[1+\gamma]{\frac{1+\gamma}{\gamma M }[f(\wb^0) - f_*]}$ minimizes the right-hand side of \eqref{ineq-to-min} and for this choice of $c_\gamma$,  the inequality \eqref{ineq-to-min} becomes

$$ \min_{0\leq k \leq K-1} \mathbb{E} \| \nabla f(\wb^k)\|^2 \leq \frac{a_\gamma}{K^{\gamma/(1+\gamma)}}
$$ 
with $a_\gamma = \sigma_\gamma \left( \sqrt[\gamma+1]{\frac{(1+\gamma)}{\gamma} M}\right) [f(\wb^0) - f_*]^{\frac{\gamma}{\gamma+1}}. $ 
\end{proof}

\bibliography{sgd_tail,work,levy}

\end{document}